\documentclass{article}

\usepackage{neurips_data_2022}

\usepackage[utf8]{inputenc} %
\usepackage[T1]{fontenc}    %
\usepackage{hyperref}       %
\usepackage{url}            %
\usepackage{booktabs}       %
\usepackage{amsfonts}       %
\usepackage{amsmath}
\usepackage{nicefrac}       %
\usepackage{microtype}      %
\usepackage{xcolor}         %
\usepackage{graphicx}
\usepackage{tikz}
\usepackage{amssymb}
\usepackage[show]{notes-alt}
\usepackage{multirow}
\usepackage{multicol}
\usepackage{xspace}
\usepackage{rotating}
\usepackage{siunitx} %
\usepackage{tcolorbox}
\usepackage{framed}
\usepackage{bbm}
\usepackage{enumitem} %

\newcommand{\fidelity}{RDT-Fidelity \xspace}
\newcommand{\mpara}[1]{\medskip\noindent{\textbf{#1}}}

\usepackage{pifont}

\usepackage{wrapfig}
\newcommand{\ourbench}{\textsc{Bagel}\xspace}
\newcommand{\XAI}{explainers \xspace}
\newcommand{\cmark}{\ding{51}}
\newcommand{\xmark}{\ding{55}}

\newcommand{\gnnexp}{GNNExp\xspace}
\newcommand{\pgm}{PGM\xspace}
\newcommand{\zorro}{Zorro\xspace}

\newcommand{\gradcam}{GradCAM\xspace}

\newcommand{\gcn}{GCN\xspace}
\newcommand{\gat}{GAT\xspace}
\newcommand{\gin}{GIN\xspace}
\newcommand{\appnp}{APPNP\xspace}

\newcommand{\cora}{\textsc{Cora}\xspace}
\newcommand{\citeseer}{\textsc{CiteSeer}\xspace}

\newcommand{\movie}{Movie Reviews\xspace}

\usepackage{tikz}
\usepackage{pgfplots}
\usepackage{subcaption}
\usepackage{natbib}
\setcitestyle{numbers}
\nolinenumbers

\title{ \ourbench: A \underline{B}enchmark for \underline{A}ssessing \underline{G}raph Neural Network \underline{E}xp\underline{l}anations}

\author{%
  Mandeep Rathee \\
  L3S Research Center, Hannover\\
  \texttt{rathee@l3s.de} \\
  
   \And
   Thorben Funke \\
    L3S Research Center, Hannover\\
   \texttt{tfunke@l3s.de} \\
   \AND
   Avishek Anand \\
  Delft University of Technology (TU Delft)\\
   \texttt{avishek.anand@tudelft.nl} \\
   \And
   Megha Khosla \\
    Delft University of Technology (TU Delft)\\
   \texttt{M.Khosla@tudelft.nl} \\
}

\begin{document}

\maketitle

\begin{abstract}
The problem of interpreting the decisions of machine learning is a well-researched and important.
We are interested in a specific type of machine learning model that deals with graph data called graph neural networks.
Evaluating interpretability approaches for graph neural networks (GNN) specifically are known to be challenging due to the lack of a commonly accepted benchmark.
Given a GNN model, several interpretability approaches exist to explain GNN models with diverse (sometimes conflicting) evaluation methodologies.
In this paper, we propose a benchmark for evaluating the explainability approaches for GNNs called \ourbench{}.
In \ourbench{}, we firstly propose four diverse GNN explanation evaluation regimes --
1) \emph{faithfulness}, 
2) \emph{sparsity}, 
3) \emph{correctness}, and 4) \emph{plausibility}. 
We reconcile multiple evaluation metrics in the existing literature and cover diverse notions for a holistic evaluation.
Our graph datasets range from citation networks, document graphs, to graphs from molecules and proteins.
We conduct an extensive empirical study on four GNN models and nine post-hoc explanation approaches for node and graph classification tasks.
We open both the benchmarks and reference implementations and make them available at \url{https://github.com/Mandeep-Rathee/Bagel-benchmark}.

\end{abstract}

\section{Introduction}
\label{sec:intro}

Graph neural networks (GNNs) \cite{velickovic2018graph,kipf2017semi,klicpera2018predict, xu2018how, hamilton2017inductive} are representation learning techniques that encode structured information into low dimensional space using a feature aggregation mechanism over the node neighborhoods. GNNs have shown state-of-the-art performance across many scientific fields in various important downstream applications, such as molecular data analysis, drug discovery, toxic molecule detection, and community clustering~\cite{dong2022mucomid,10.1093/bib/bbab159,ying2018graph}. 

 There have been benchmarks and datasets for interpretability of machine learning models~\cite{wiegreffe2021teach,liu2021synthetic}. The rising number of applications of GNNs in several sensitive domains like medicine and healthcare \cite{dong2022mucomid,lu2021weighted} necessitates the need to explain their decision-making process. GNNs are inherently black-box and non-interpretable. Moreover, due to the complex interplay of node features and neighborhood structure in the decision-making process, general explanation approaches \cite{lundberg2017unified:shap,ribeiro2016:lime,singh2020model} cannot be trivially applied for graph models. 
Consequently, several explanation techniques \cite{ying2019:gnnexplainer,funke2021zorro,vu2020pgm, XGNN2020} have been proposed for GNNs in the last few years.
Unlike standard machine learning models, in GNNs, the explanation methods return a combination of \textit{feature} and \textit{data attributions} as explanations given a trained GNN model. 
For example, in a node classification task, an explanation of a class prediction for a certain node could be a subset of the responsible features in addition to a subset of responsible nodes/edges in its computational graph. 
A known challenge in developing explanation techniques is that of evaluation of the quality of explanations.
This challenge also extends to the evaluation of explainability approaches for GNNs and is the focus of this work.

Existing approaches usually focus on a certain aspect of evaluation, sometimes even performed on synthetic datasets. For example, some works employ synthetic datasets with an already-known subgraph (sometimes referred to as the ground truth reason or simply the ground truth). 
Explanations are then evaluated based on their agreement with the ground truth. 
Such an evaluation is sometimes flawed as one cannot always guarantee that the GNN has used in the first place the seeded subgraph for its decision-making process \cite{faber2021comparing}. 
Besides, there is no standardized procedure for comparing different GNN explanations. For example, feature attribution methods can generate soft masks (feature importance as a distribution) or hard masks (boolean selections) over features. Comparing hard and soft mask explanations needs a common and standardized protocol.
Finally, the check for \textit{human plausibility} and correctness have been ignored in the evaluation of GNN explainers.
Human plausibility checks if a model predicts \textit{right for the right reason.} On the other hand, the correctness of an explanation checks if the \XAI is able to isolate spurious correlations and biases that are intentionally added to the training data as a proxy for biases present in real-world data.

\begin{figure}[!t]
    \centering
    \subfloat{{\includegraphics[width=0.75\columnwidth]{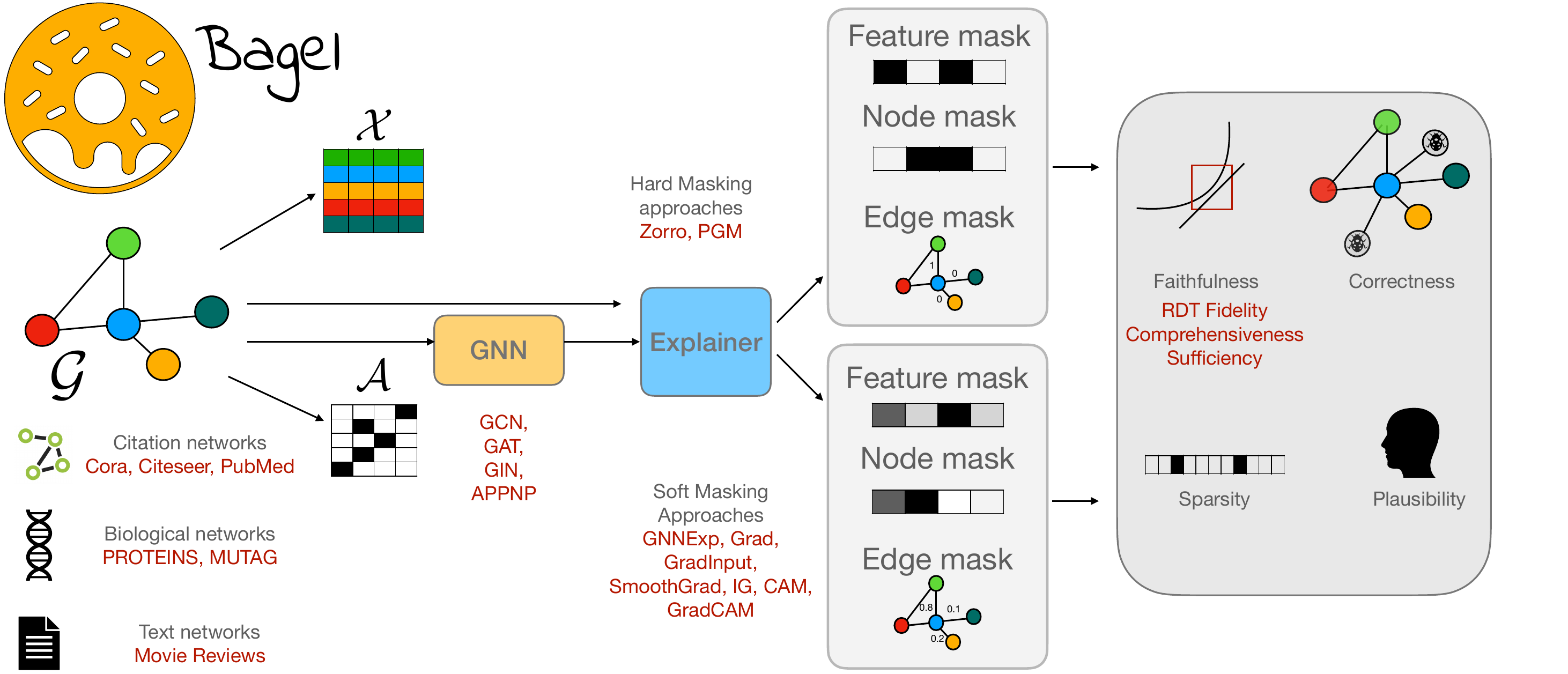} }}
    \caption{An overview of the \ourbench{} benchmark.}
    \label{fig:overview}
\end{figure}

To address the issues of a holistic evaluation and contribute a resource to the growing community on GNN explainability, we developed \ourbench{}, a benchmark platform for evaluating explanation approaches for graph neural networks or GNNs. \ourbench{} as depicted in Figure \ref{fig:overview} covers diverse datasets~\cite{sen2008collective,debnath1991structure,borgwardt2005protein,zaidan2008modeling} from the literature, a range of standardized metrics, and a modular, extendable framework for execution and evaluation of GNN explanation approaches, along with initial implementations of recent popular explanation methods. \ourbench{} includes:

\begin{itemize}[leftmargin=*]
    \item [$\circ$]  Four diverse evaluation notions that evaluate the \textit{faithfulness, sparsity, correctness, and plausibility} of GNN explanations on real-world datasets.
    While the first three metrics focus on evaluating the explainers, plausibility checks for explanations to be human congruent.
    
    \item [$\circ$] Besides the widely used datasets for measuring faithfulness of explanations, \ourbench{} consists of new datasets for the plausibility of explanation approaches in our benchmark datasets.

    \item [$\circ$] We unify multiple evaluations, metrics, domains, and datasets into an easy-to-use format that reduces the entry barrier for evaluating new approaches to explain GNNs. Additionally, we provide an extendable library to implement and evaluate GNN explainers.

    \item [$\circ$] We conduct an extensive evaluation of GNNExplainer(\gnnexp)~\cite{ying2019:gnnexplainer}, PGM-Explainer(PGM)~\cite{vu2020pgm}, Zorro~\cite{funke2021zorro}, Grad~\cite{imageclasssimonyan2013deep}, GradInput\cite{imageclasssimonyan2013deep}, Integrated Gradient(IG)~\cite{sundararajan2017:integratedgrad}, SmoothGrad~\cite{smilkov2017smoothgrad}, CAM~\cite{pope2019explainability} and GradCAM~\cite{pope2019explainability} existing GNN \XAI in \ourbench{}.

\end{itemize}

We show that there is no clear winner in GNN explanation methods showing nuanced interpretations of the GNN explanation methods using the multiple metrics considered. We finally note that evaluating the effectiveness of explanations is an intrinsically human-centric task that ideally requires human studies. 
However, the goal of \ourbench{} is to provide a fast and accurate evaluation strategy that is often desirable to develop new explainability techniques using empirical evaluation metrics before the human trial stage. The code and the datasets used in our benchmark are available at \linebreak \url{https://github.com/Mandeep-Rathee/Bagel-benchmark}.
 
\section{Background and Preliminaries}
\label{sec:Related Work}
\mpara{Graph Neural Networks.} Let $\mathcal{G(V,E)}$ be a graph with $\mathcal{V}$ is a set of nodes and $\mathcal{E}$ is a set of edges. Let $\mathcal{A}\in\{0,1\}^{(n,n)}$  be the adjacency matrix of the graph where $n$ is the number of nodes in the graph with $\mathcal{A}_{ij}=1$ if there is an edge between node $i$ and $j$ and $0$ otherwise. Let $\mathcal{X}\in\mathbf{R}^{(n,d)}$ be the features matrix where $d$ is the number of features. For a given node $v\in \mathcal{V}$, $\mathbf{x}_{v}$ denotes its features vector and $\mathcal{N}_{v}$ is a set of its neighbors. We denote the trained GNN model as $f$ on the given graph. For each layer $\ell$, the representation of node $v$ is obtained by aggregating and transforming the representations of its neighboring nodes at layer $\ell-1$
\begin{align}
    \boldsymbol{h}_{v}^{(\ell)}=&\operatorname{AGG}\left(\left\{\mathbf{x}_{v}^{(\ell-1)},\left\{\mathbf{x}_{u}^{(\ell-1)} \mid u \in{\mathcal{N}}_{v}\right\}\right\}\right), \quad
    \mathbf{x}_{v}^{(\ell)}= \operatorname{TRANSFORM}
    \left(\boldsymbol{h}_{v}^{(\ell)}, W^{(\ell)}\right),
\end{align}
where $W^{(\ell)}$ represents the weight matrix at layer $\ell$. 
The aggregation function $\operatorname{AGG}$ function is dependent on the GNN model. For example, graph convolution network (GCN)~\cite{kipf2017semi} uses a degree weighted aggregation of neighborhood features, whereas graph attention network (GAT)~\cite{velickovic2018graph} learns neighborhood weights via an attention mechanism. The prediction can be obtained at the final layer using a \emph{softmax function}. An additional pooling layer is applied for \textit{graph classification} tasks before applying \emph{softmax function}.

\subsection{Post-hoc explanations and evaluation for GNNs} 
Post-hoc explainers for GNNs produce feature and local structure attributions where a combination of a masked set of nodes, edges, and features is retrieved as an explanation. To compute the explanation for a $k$-layer GNN, the $k$-hop neighborhood (also referred to as the node's computational graph) of the node is utilized. \ourbench currently consists of 3 classes of post-explanation techniques: \emph{gradient based}, \emph{perturbation based} and \emph{surrogate model} approaches. 
The gradient-based methods~\cite{imageclasssimonyan2013deep, sundararajan2017:integratedgrad,smilkov2017smoothgrad,pope2019explainability} 
are the simplest approaches for generating the explanation for any differentiable trained model. 
In these approaches, the importance scores for the explanation are usually computed using gradients of the input. The perturbation-based approaches~\cite{funke2021zorro,ying2019:gnnexplainer,luo2020parameterized,yuan2021explainability,schlichtkrull2020interpreting} learns the important features and structural information by observing the predictive power of the model when noise is added to the input. The surrogate-based approaches~\cite{vu2020pgm,huang2020graphlime,zhang2021relex} learns a simple interpretable model for the local neighborhood of a query node and its prediction. 
The explanations generated by this simple model are treated as the explanations of the original model. We note that \ourbench is, in general, applicable for any explainer which returns binary (hard) or continuous (soft) importance scores (as depicted in Figure \ref{fig:overview}) for the input features/nodes/edges as an explanation.

\subsection{Related work on evaluation for post-hoc explanations}

Evaluation of explanation methods for any predictive model is inherently tricky.
Specifically, when evaluating already trained models, we are faced with the \textit{lack of true explanations}. Collecting true explanations (sometimes referred to as ground truth) for GNNs is even more challenging due to the abstract nature of the input graphs. Moreover, depending on the explanation collection method, it is not always clear if the model used the ground truth in its decision-making process.

Nevertheless, some current works employ small synthetic datasets seeded with a ground truth subgraph. Consequently, metrics such as \textit{explanation accuracy} \cite{sanchez2020evaluating,ying2019:gnnexplainer} were proposed, which measure the agreement of found explanation with that of ground truth. Observing the false optimism of accuracy metric for small explanation subgraphs \cite{funke2021zorro} proposed the use of \textit{Precision} instead of accuracy.

Another important notion is \emph{faithfulness} where the key idea is to measure how much the explanation characterizes the model's working. To measure faithfulness \cite{sanchez2020evaluating} degrade model performance by damaging the training dataset and measuring how each explanation method responds. 
The lack of ground truth again limits such a measure. \cite{pope2019explainability} proposed to compute faithfulness as the difference of accuracy (or predicted probability) between the original predictions and the new predictions after masking out the input features found by the explanation. This was called \textit{Fidelity} in their work. As the features cannot be removed in entirety to measure their impact \cite{funke2021zorro} proposed  \fidelity based on rate distortion theory defined as the expected predictive score of an explanation over all possible configurations of the non-explanation features.

An important criterion to measure the goodness of an explanation is its size. For example, the full input is also a faithful explanation. However, humans find shorter explanations easier to analyze and reason. Works such as \cite{pope2019explainability} measure the sparsity of an explanation as the fraction of features selected by the explainer. Noting that this definition is not directly applicable for softmask approaches, \cite{funke2021zorro} proposes to quantify sparsity as entropy over the normalized distribution of explanation masks.

The authors in \cite{sanchez2020evaluating} argued that the explanation should be stable under input perturbations. In particular, for graph classification, they perturbed test graphs by adding a few nodes/edges such that the final prediction remains the same as that for an unperturbed graph. Lower the change in explanation under perturbations better the stability. A challenge here is that there is no principled way to find the perturbations. For example, a part of the explanation might be altered under random perturbations even if the prediction is unchanged.

\section{\ourbench: A Unified Framework for Evaluating Explanations}
We now present our framework  \ourbench for evaluating GNN explanations. Specifically, \ourbench unifies existing and our proposed notions into 2 main classes. In the \emph{first class} of measures we aim at evaluating the explanation methods in the sense that whether they are truly describing the model's workings. The first category includes three metrics: \emph{faithfulness}, \emph{sparsity} and \emph{correctness}. Faithfulness determines if an explanation alone can replicate the model's behavior. Sparsity focuses on rewarding shorter explanations. Correctness determines if the explanation model is able to detect any injected correlations responsible for altering model's behavior. 
The metrics in the second class are aimed at evaluating the GNN model itself. Here we propose \emph{plausibility} which measures how close is the decision making process of the trained model (as revealed by explanations) to human rationales.

\subsection{Faithfulness: Can explanations approximate model's behavior?}
\label{subsubsec:Faithfulness} The key idea here to evaluate the ability of the explanation to characterize model's working. Unlike previous works we argue that there is not a single measure for faithfulness which can be effectively used for all kinds of datasets and explanations. Consequently we propose a set of two measures to quantify faithfulness depending on the dataset/explanation type. 

\begin{itemize}[leftmargin=*]
\item \textsc{Rate distortion based fidelity.}
The fidelity of an explanation is usually measured by the ability of an explanation to  approximate the model behavior \cite{ribeiro2016:lime}.
For explanations which contain the feature attributions with or without structure attributions,
we propose the rate distortion theory based metric proposed in \cite{funke2021zorro} to measure the fidelity of an explanation.
In short, a subgraph of the node's computational graph and its set of features
are relevant for a classification decision if the expected classifier score remains nearly the same when randomizing the remaining features.

The \textit{\fidelity{}} of explanation $\mathcal{S}$ corresponding to explanation mask $M(\mathcal{S})$ with respect to the GNN $f$, input $X$ and the noise distribution $\mathcal{N}$ is given by
\begin{equation}
\label{eq:fidelity}
\mathcal{F}(\mathcal{S}) = \mathbb{E}_{Y_{\mathcal{S}}|Z\sim \mathcal{N}} \left[\mathbbm{1}_{f\left(X\right)=f(Y_{\mathcal{S}})}\right].
\end{equation}
where the perturbed input is given by
\begin{equation}
 Y_{\mathcal{S}} = X\odot M(\mathcal{S}) + Z\odot(\mathbbm{1} - M(\mathcal{S})), Z\sim \mathcal{N},
    \label{eq:noisy_features_matrix}
\end{equation}
where $\odot$ denotes an element-wise multiplication, and $\mathbbm{1}$ a matrix of ones with the corresponding size and $\mathcal{N}$ is a noise distribution. We choose the noise distribution to the global empirical distribution of the features. This makes sure that the perturbed data points are still in the same distribution as the original data ~\cite{hooker2019benchmark}.

\mpara{Connection to explanation stability.} As shown in \cite{funke2021zorro} explanations with high RDT-fidelity are highly stable. High fidelity score implies that the explanation has high predictive power under perturbations of the rest of the input. Unlike the strategy of \cite{sanchez2020evaluating} to evaluate explanation stability, it is here ensured that the explanation itself is never altered.

\mpara{The special case of dense feature representations.} For some datasets it is more appropriate to consider only structure based explanations. For example, when features themselves are dense representations extracted using some black-box embedding method, feature explanations as well as feature perturbations might not make much sense. It is then more appropriate to check the abilities of the explanation with the rest of nodes/edges removed and keeping the features intact. Towards that we employ the following measures of comprehensiveness and sufficiency also used in \cite{deyoung2019:eraser}.

\item \textsc{Comprehensiveness and Sufficiency.} For explanations which contain only nodes or/and edges we adapt the comprehensiveness and sufficiency measures of \cite{deyoung2019:eraser} for GNNs.  
Let $\mathcal{G}$ is the graph and $\mathcal{G^\prime}\subseteq \mathcal{G}$ is the explanation graph with important (attribution) nodes/edges. Let $f$ be the trained GNN model and $f(\mathcal{G})_j$ be the prediction made by GNN for $j^{th}$ class. We measure fidelity by \textit{comprehensiveness} (which answers the question if  all nodes/edges in the graph needed to make a prediction were selected?) and \textit{sufficiency} (if the extracted nodes/edges are sufficient to come up the original prediction?)
\begin{equation}
\textit{sufficiency}=f\left(\mathcal{G}\right)_{j}-f\left(\mathcal{G^\prime}\right)_{j}, \quad \textit{comprehensiveness}=f\left(\mathcal{G}\right)_{j}-f\left(\mathcal{G} \backslash \mathcal{G^\prime}\right)_{j}
\end{equation}

The high \textit{comprehensiveness} value shows that the prediction is most likely because of the explanation $\mathcal{G^\prime}$ and low \textit{comprehensiveness} value shows that $\mathcal{G^\prime}$ is mostly not responsible for the prediction. Since most of the explainers retrieve soft masks we employ aggregated \textit{comprehensiveness} and \textit{sufficiency} measures. In particular, we divide the soft masks into $|\mathcal{B}|=5$ bins by using top $k\in \mathcal{B} = \{1\%, 5\%, 10\%, 20\%, 50\% \}$ of the explanation with respect to the soft masks values~\cite{samek2016evaluating}. The aggregated \textit{sufficiency} is defined as:
$\frac{1}{|\mathcal{B}|}\left(\sum_{k=1}^{|\mathcal{B}|} f\left(\mathcal{G}\right)_{j}-f\left(\mathcal{G^\prime}_{k}\right)_{j}\right).$ The aggregated \textit{comprehensiveness} is defined in similar fashion.

\end{itemize}

\subsection{Sparsity: Are the explanations non trivial?}
\label{sub:Sparsity}
High faithfulness ensures that the explanation approximates the model behavior well. However, the complete input completely determines the model behavior. Thus explanation sparsity is an important criteria for evaluation.
Let $p$ be the normalized distribution of explanation (feature) masks. Then sparsity of an explanation is given by the entropy $H(p)$ and is bounded from above by $\log(|M|)$ where $M$ corresponds to a complete set of features or nodes. While an entire input can be a faithful explanation it is important to evaluate an explanation with respect to its size. A shorter explanation is easier to analyse and is more human understandable. We adopt the entropy based definition of sparsity as in \cite{funke2021zorro} because of its applicability to both soft and hard explanation masks. 
In particular, let $p$ denote the normalized distribution of node/edge/feature masks. We compute sparsity of an explanation as the entropy over the mask distribution: 
$H(p)= -\sum_{\phi\in M} p(\phi) \log p(\phi).$ 

\subsection{Correctness: Can the explanations detect externally injected correlations?}
\label{subsubsec:Correctness}
While the above measures are essential that the given explanation is predictive certain applications might need explanations for model debugging, for example to detect any spurious correlations picked up by model thereby increasing model bias. Towards that we measure correctness of an explanation in terms of its ability to recognize the \textit{externally injected correlations} which alters the model decision. A switch in model decision is an evidence of the use of these injected correlations in the actual decision making process. 

In particular, we first choose a set of  incorrectly labelled nodes, $V$. To each such node $v$, we add edges to the nodes in the training data which have the same label as $v$. We call such edges \textit{decoys}. We retrain the GNN model with the perturbed data. We measure the correctness of explanation $\mathcal{S}$ for nodes in $V$ which are now correctly predicted in terms of precision and recall of the decoys in the returned explanation:
$
    Precision_C = {N_{de} \over N_e}, ~~~~ Recall_C ={N_{de} \over N_d}, 
$
where $N_{de}$ is the number of decoys in the obtained explanation, $N_{d}$ total number of decoys injected and $N_{e}$ is the size of the retrieved explanation. Note that our proposed approach of using injected correlations is different from using a synthetic graph with seeded ground truth. In particular, for seeded graph approach it is not always clear if the ground truth is actually picked up by the model to make its decision.

\subsection{Plausibility: How close is the model's decision process to humans rationals?} 
Human congruence or plausibility \cite{lei2016:rationales,lage2019evaluation,strout2019human} tries to establish how close or congruent is the trained model to human rationales for solving a predictive task.
Trained models often exhibit the \textit{clever-hans effect}, that is predictive models can adopt spurious correlations in the training data or due to misplaced inductive biases that have right results for the wrong reasons.
Towards this, data is collected from humans for perceptive tasks where humans explicitly provide their rationales.
These human rationales are in used as ground truth for evaluating if trained models are right for the right reasons.

For applications where obtaining human rationales is indeed possible we propose the use of token-level F1  for binary explanation masks and area under precision recall curve (AUPRC) for soft masks. The reader might have noticed that this metric is similar to the explanation accuracy in earlier works. We argue against the use of term \textit{accurate} to measure plausibility as similarity to human rationale does not always guarantee that the model has learnt an explanation which contains the reasoning of the model itself and not only of the humans.

\section{Experimental Setup}
\vspace{-0.1cm}

\mpara{Models and Explainers.} We demonstrate the use and advantage of the proposed framework by evaluating $9$ explanation methods over $6$ datasets and $4$ GNN models.  Currently our benchmark consists of these GNN models: 
graph convolutional networks (\gcn)~\cite{kipf2017semi}, graph attention network (\gat)~\citep{velickovic2018graph}, the approximation of personalized propagation of neural predictions (\appnp)~\citep{klicpera2018predict}, and graph isomorphism network (\gin)~\citep{xu2018how}. The models were chosen based on their differences in (i) exploiting inductive biases (based on different feature aggregation strategies), (ii) test performance (see tables \ref{tab:dataset stat for node classification} and \ref{tab:dataset stat for graph classification} in the Appendix) and (iii) response to injected correlations (see Table \ref{tab:cora_citeseer_correctness_summary} and the corresponding discussion). 

We perform experiments with perturbation based approaches like GNNExplainer (\gnnexp)~\cite{ying2019:gnnexplainer} and Zorro~\cite{funke2021zorro}, surrogate methods like PGM-Explainer (PGM)~\cite{vu2020pgm}, and gradient-based approaches like Grad~\cite{imageclasssimonyan2013deep}, GradInput\cite{imageclasssimonyan2013deep}, Integrated Gradient (IG)~\cite{sundararajan2017:integratedgrad}, SmoothGrad~\cite{smilkov2017smoothgrad}, CAM~\cite{pope2019explainability} and GradCAM~\cite{pope2019explainability}. \gnnexp~returns soft feature masks and edge masks. We transform the edge masks into node masks, in which we equally distribute the edge importance score to both nodes sharing the edge. As already mentioned \ourbench is extendable, and more approaches and explainers can be easily added. 

\subsection{Datasets}
 We now describe new and existing datasets used in our evaluation framework and the corresponding rationale. 

\textbf{New Dataset for Plausibility.} To measure the plausibility of an explanation, we first require the corresponding human rationales. Since the existing graph datasets do not have such annotated information, we transform a text sentiment prediction task into a graph classification task. Specifically, we adopt the Movie Reviews dataset~\cite{zaidan2008modeling} from the ERASER benchmark~\citep{deyoung2019:eraser}.
The task is the binary classification, which is the differentiation between positive and negative movie reviews. A detailed description of dataset construction is given in Appendix~\ref{sec:datasets}.

\textbf{Dataset to measure Comprehensiveness and Sufficiency.} Note that the measures comprehensiveness and sufficiency are only applicable for node/edge explanations. We use the above-described Movie Review dataset for these two measures too. The rationale is that the node features are generated using Glove and are not human-understandable. In this case, a structure-based explanation would be more meaningful than a feature-based one.

\textbf{Datasets for Correctness.}
We employ two citation datasets \textsc{Cora}~\cite{sen2008collective} and \textsc{CiteSeer}~\cite{sen2008collective}. 
After injecting correlations/decoys corresponding to incorrectly labeled nodes as described in Section~\ref{subsubsec:Correctness}, we re-train the GNN model. The rationale behind adding homophily increasing correlations is the observation from previous works~\cite{khosla19comparative,zhu2020beyond} that GNN's performance increases with higher homophily. A model will have picked up these correlations if the previous incorrect nodes are now correctly predicted. We further evaluate the correctness of the explanation only for newly correctly predicted nodes. 

\textbf{Datasets for RDT-Fidelity.}
\label{subsec:fidelity_ex_setup}
We perform the \textit{RTD-Fidelity} evaluation on both \textit{node classification} and \textit{graph classification} tasks. At node level, we use Citation datasets. Table~\ref{tab:dataset stat for node classification} shows the dataset statistics and GNNs performances. 
For node classification task, we select 300 nodes for \textsc{Cora} and \textsc{CiteSeer} and 700 nodes for \textsc{PubMed} randomly. For graph classification task, we select 50 graphs for both MUTAG~\cite{debnath1991structure} and PROTEINS~\cite{borgwardt2005protein} datasets. 

\section{Result Analysis}

\subsection{Faithfulness}
\label{subsec:Faithfulness_result}
\begin{table*}[h!!]
    \centering
    \caption{Results for \textit{RDT-Fidelity} for node classification.}
    {\small
    \setlength{\tabcolsep}{3.0pt}
    \begin{tabular}{ll ccc ccc ccc ccc}
\toprule
\multirow{2}{*}{\small {Mask}} & \multirow{2}{*}{Methods} & \multicolumn{4}{c}{\textsc{Cora}} & \multicolumn{4}{c}{\textsc{CiteSeer}} & \multicolumn{4}{c}{\textsc{PubMed}} \\
 &  &  \small GCN &\small   GAT & \small  GIN & \small APPNP & \small GCN & \small  GAT & \small  GIN & \small APPNP & \small GCN & \small GAT & \small GIN & \small APPNP \\
\midrule
\multirow{2}{*}{Hard} & Zorro    &  \textbf{0.97} &  \textbf{0.97} &  \textbf{0.96} &  \textbf{0.97} & \textbf{0.97} &  \textbf{0.97} &  \textbf{0.97} &  \bf{0.96} &   \bf{0.96} &  \bf{0.97} &  \bf{0.97} & \bf{0.96} \\
 & PGM   & 0.84 &  0.77 &  0.60 &  0.89 & 0.92 & 0.93 &  0.73 &  0.95 &   0.78 &  0.69 &  0.74 &  0.96 \\ 
\midrule 
\multirow{6}{*}{Soft} & \gnnexp   &  0.71 &  0.66 &  0.52 &  0.65 &     0.68 &  0.69 &  0.51 &  0.62 &   0.67 &  0.73 &  0.67 &  0.72 \\
 & Grad           &  0.15 &  0.18 &  0.19 &  0.17 &     0.17 &  0.19 &  0.28 &  0.18 &   0.37 &  0.43 &  0.42 &  0.37 \\
 & GradInput      &  0.15 &  0.18 &  0.18 &  0.16 &     0.16 &  0.18 &  0.26 &  0.17 &   0.36 &  0.42 &  0.42 &  0.36 \\
 & SmoothGrad     &0.44 &  0.42 & 0.38 & 0.50 &  0.54 &  0.57 &  0.45 & 0.62  &  0.52 &  0.53& 0.67 & 0.59 \\ 
  & IG      & 0.45 & 0.47 & 0.26 & 0.51 &  0.53 &  0.70 &  0.45 & 0.62 & 0.52 & 0.56 &  0.68 & 0.59 \\

 & Empty Expl.      &  0.15 &  0.18 &  0.18 &  0.16 &     0.16 &  0.18 &  0.26 &  0.17 &   0.36 &  0.42 &  0.42 &  0.36 \\
\bottomrule
\end{tabular}
    }
    \label{tab:node_fidelity}
\end{table*}

\textbf{RDT-Fidelity.}
In Table~\ref{tab:node_fidelity} we compare the RDT-Fidelity scores of various explanation methods. As a trivial baseline, we include an empty explanation baseline, i.e., all features are set randomly using the noise distribution. A common feature of \zorro and \pgm is that they both learn the explanations from a sampled local dataset. The local dataset is created by perturbing the features of nodes from the computational graph (neighborhood of query nodes). While they employ different optimization strategies to find explanations, the result is a stable explanation that also reflects our results. The gradient-based explanations achieve the lowest fidelity.

For the graph classification task (see Table \ref{tab:graph_fidelity} in Appendix), all methods, including the gradient-based approaches, perform relatively well except for the \gcn model. \pgm shows more consistent performance across all models and datasets. We leave out \zorro as it is not applicable for the graph classification task.

\begin{wraptable}{r}{9cm}
  \vspace{-5.5mm}
    \caption{Faithfulness as \textit{comprehensiveness} and \textit{sufficiency} measured for \movie dataset. Low sufficiency and high comprehensiveness indicates high faithfulness.}
    {\small
    \renewcommand*{\arraystretch}{0.5}
    \setlength{\tabcolsep}{2pt}
    \begin{tabular}{ll cc cc cc cc }
\toprule
{\textbf{}} &  \multirow{2}{*}{Models}   
 &  \multicolumn{2}{c}{GCN} &   \multicolumn{2}{c}{GAT} &  \multicolumn{2}{c}{GIN} & \multicolumn{2}{c}{APPNP}   \\
\small
&  &Suff. & Comp. &Suff.&Comp. &Suff.& Comp. &Suff. &Comp. \\
\midrule
 &\gnnexp  & 0.56 &-0.01 & 0.47 &0.03  & 0.24 &0.32 & 0.48 &0.07  \\
 & Grad          & \textbf{0.02}&  \textbf{0.39}  & 0.15& 0.16 & 0.25 &0.31  & 0.11 &0.20 \\
 & GradInput     & 0.07 &0.36 & 0.14& 0.16  &0.22 & \textbf{0.33}  & 0.12& 0.24  \\
  & SmoothGrad & 0.08 & 0.34 & 0.15& 0.23  & 0.25& 0.27  & 0.14& 0.23  \\
 & IG    &  0.11& 0.38  & 0.16& 0.21 & 0.23& 0.27 & 0.16& 0.24 \\
 & CAM    &  0.41& 0.01 & 0.11 &0.14 & 0.26 &0.26 & 0.39& 0.05  \\
 & GradCAM    &  \textbf{0.02} & 0.29  & \textbf{0.06}& \textbf{0.27} & \textbf{0.21}& 0.27  & \textbf{0.07} &\textbf{0.28}  \\
\bottomrule
\end{tabular}
    }
    \vspace{-4.5mm}
    \label{tab:suff and comp}
\end{wraptable}

\textbf{Comprehensiveness and Sufficiency.} We evaluate faithfulness for explanations for the \movie dataset using aggregated comprehensiveness and sufficiency measures. The results for soft-mask explanations are shown in Table~\ref{tab:suff and comp}. \gradcam has the lowest \textit{sufficiency} which suggests that the explanations are sufficient to mimic the prediction of GNN models. On the other hand, the explanations generated by \gradcam with \gcn and \gin suggest that there still exists important part of the input outside of the explanations which are required to approximate the GNN's prediction. On the other hand, \gnnexp, which so far outperformed gradient-based explanations for the node classification task, shows the worst sufficiency and comprehensiveness. Even if we use the complete feature set and only the node masks to evaluate explanations, the node masks for \gnnexp are learned together with the feature explanations. This differs from gradient-based approaches, which ignore feature and structure explanation trade-offs. The current performance of \gnnexp indicates that it might not be appropriate to use entangled features and structure explanations independently.

\subsection{Sparsity}
As already mentioned, a complete input could already do well on all faithfulness measures. Therefore, we further look for sparser explanations. The results for node sparsity for explanations in node classification task are provided in Table \ref{tab:node_sparsity}. 
 For the hard masking approaches (Zorro and PGM), Zorro outperforms PGM with all GNN models except for GIN. Conversely, there is no clear winner for the soft mask approach. The high sparsity for soft-masking approaches implies a near-uniform node attribution and consequently lower interpretability. In general, faithfulness and sparsity of an explanation should be analyzed together. A uniformly distributed explanation mask could already provide an explanation with high faithfulness as it leads to using the complete input as an explanation.

\begin{table*}[htbp]
    \centering
    \caption{Results for sparsity (computed as entropy over mask distribution) for node classification. The lower the score sparser is the explanation.}
    {\small
    \setlength{\tabcolsep}{3.0pt}
    \begin{tabular}{ll ccc ccc ccc ccc}
\toprule
\multirow{2}{*}{\small {Mask}} & \multirow{2}{*}{Methods} & \multicolumn{4}{c}{\textsc{Cora}} & \multicolumn{4}{c}{\textsc{CiteSeer}} & \multicolumn{4}{c}{\textsc{PubMed}} \\
 &  &  \small GCN &\small   GAT & \small  GIN & \small APPNP & \small GCN & \small  GAT & \small  GIN & \small APPNP & \small GCN & \small GAT & \small GIN & \small APPNP \\
\midrule
\multirow{2}{*}{Hard} & Zorro  &  \textbf{1.58} &  \textbf{1.59} &  2.17 &  \textbf{1.48} &     \textbf{1.26} &  \textbf{1.09} &  1.58 &  \textbf{1.07} &   \textbf{1.51} &  \textbf{1.31} &  2.18 &  \textbf{1.25} \\
  & PGM    &  2.06 & 1.82 &  \bf{1.66} & 1.99 & 1.47 &  1.59 & \bf{1.10} & 1.54 & 1.64 &  1.16 & \bf{1.62} &  2.93 \\
\midrule
\multirow{5}{*}{Soft} & \gnnexp   &   2.48 &  2.49 &  2.56 &  2.51 &     1.67 &  1.67 &  1.70 &  1.68 &    2.70 &  2.71 &  2.71 &  2.71 \\
 & Grad            &  2.48 &  2.34 &  2.25 &  2.35 &     1.70 &  1.61 &  1.55 &  1.60 &   2.91 &  2.76 &  3.11 &  2.73 \\
 & GradInput        &  2.53 &  2.43 &  2.23 &  2.41 &     1.61 &  1.58 &  1.54 &  1.52 &   3.02 &  2.94 &  3.41 &  2.81 \\
&SmoothGrad &2.48  &2.52  &2.91  &2.31  &1.77 &1.77 &1.93  &1.66  &2.89  &3.02   &3.23  &2.54   \\  
& IG & 2.49 &2.50  & 2.84 &2.31  &1.76 &1.77 &1.91  &1.66  &2.84  &2.89   &3.06  &2.58   \\ 

\bottomrule
\end{tabular}
\vspace{-0.5cm}
    }
    \label{tab:node_sparsity}
\end{table*}

\subsection{Correctness}
\label{subsec:Correctness_result}

\begin{wraptable}{r}{6cm}
    \small
    \vspace{-6mm}
    \caption{The number of incorrectly labelled nodes (\xmark) decreases after addition of decoys. The number of new correctly labelled nodes after injecting decoys is listed under \cmark.}
    \setlength{\tabcolsep}{4.5pt}
    \begin{tabular}{ll ccc ccc}
        \toprule
        \multirow{2}{*}{\textsc{Model}} &  \multicolumn{3}{c}{\textsc{Cora}} &  \multicolumn{3}{c}{\textsc{CiteSeer}} \\ 
        &  \xmark  & \cmark   & $\uparrow$(\%)   & \xmark  & \cmark   & $\uparrow$(\%)  \\
        \midrule
        GCN   & 88 & 79 &89.7 & 329  & 229 & 69.6 \\
        GAT   & 86 & 85 & 98.8 & 311  & 301 & 96.7 \\
        GIN   & 6 & 6 & 100 & 56  & 56 & 100 \\
        APPNP  & 73 & 70 & 95.8 & 280  &252 &90.0 \\
        \bottomrule
    \end{tabular}
    \label{tab:cora_citeseer_correctness_summary}
    \vspace{-5mm}
\end{wraptable}
The correctness results corresponding to different models and explainers are reported for \cora (in
Table~\ref{tab:node_correctness_cora}) and \citeseer (in Table~\ref{tab:node_correctness_citeseer}). We report precision, recall and F1 score by choosing top k nodes for the soft explanations. For hard masked approaches the number of returned nodes is listed under $|\mathcal{S}|$. Note that the number of decoys added per node is $10$. For Table~\ref{tab:node_correctness_cora} and Table~\ref{tab:node_correctness_citeseer} we use $k=20$.

In Table~\ref{tab:cora_citeseer_correctness_summary}, the effect of decoys can be seen where most of the earlier incorrectly classified nodes are now correctly classified except for GCN on \textsc{CiteSeer}.
 We also observe that number of selected nodes for GIN is very low for \textsc{Cora} dataset (i.e., only a few nodes were initially incorrectly labeled). 
\gnnexp~outperforms all other based explainers in detecting the injected correlations for both \textsc{Cora} and \textsc{CiteSeer} (detailed results moved to Table~\ref{tab:node_correctness_citeseer} in the Appendix due to space constraints). 

Comparing soft mask and hard mask approaches in this setting is tricky as for some approaches like \zorro, we cannot control the explanation size. For example, for \gat \zorro retrieved an explanation of size 40. A precision of 0.25 shows that it found all 10 injected correlations. Lack of feature ranking, as in soft mask approaches, makes it difficult to evaluate hard mask approaches for Correctness. 
For fairer evaluation, we further plot the performance of soft mask approaches with different $k$ in Appendix~\ref{sec: Why evaluation with soft masks is difficult?}. For example, the \gnnexp~shows high improvement when we increase the size of the explanation to 15. It is not surprising to see the performance degrades when we increase the size of the explanation further since it already had captured all injected decoys. Now it returns some irrelevant nodes in the explanation. Furthermore, in Table~\ref{tab:node_correctness_cora_mean} and \ref{tab:node_correctness_citeseer_mean}, we use mean as a threshold to generate hard masks. As the mean threshold turns out to be very low for all approaches, almost all nodes of the computational graph are selected as the explanation. Consequently, we observe a very low correctness score (when measured in terms of precision).

\begin{table*}[htbp]
    \centering
    \caption{Correctness  of the explanation for node classification on \textsc{Cora} dataset. We use $k=20$.}
    {\small
    \setlength{\tabcolsep}{2.3pt}
    \begin{tabular}{ll cccc cccc cccc cccc}
\toprule
\multirow{3}{*}{Mask}&\multirow{3}{*}{Methods} &  \multicolumn{16}{c}{\textsc{Cora}} \\ 
&  &\multicolumn{4}{c}{GCN} & \multicolumn{4}{c}{GAT}  &\multicolumn{4}{c}{GIN} & \multicolumn{4}{c}{APPNP} \\ 
& & P@k & R@k &F1 & |$\mathcal{S}$|& P@k & R@k &F1 & |$\mathcal{S}$| & P@k & R@k &F1 & |$\mathcal{S}$| & P@k & R@k &F1 & |$\mathcal{S}$|\\
\midrule
\multirow{2}{*}{Hard} & \zorro    &0.19 & 0.80 & 0.30 & 45 & 0.25 &0.83  & 0.37& 40 & 0.26 & 0.45 &0.27 & 33 &  0.22 & 0.79 & 0.33 & 38   \\
  & PGM      & 0.11&0.22& 0.15 & 20 &  0.18&0.36&0.24 & 20 &  0.18&0.36& 0.25 & 20 & 0.19&0.38& 0.25&20   \\
\midrule 
\multirow{5}{*}{Soft} & \gnnexp   &  \textbf{0.42}&\textbf{0.84}&\textbf{0.56} &20 &  \textbf{0.44}&\textbf{0.88}&\textbf{0.59} &20 & \textbf{0.50}&\textbf{1.00}&\textbf{0.67} &20 &  \textbf{0.34}&\textbf{0.67}& \textbf{0.58} &20  \\
 & Grad   & 0.23&0.46 &0.31  & 20 &0.29&0.58 & 0.39 & 20 & 0.30&0.60 & 0.40 &20 & 0.33&0.67 &0.45 &20  \\
 & GradInput  & 0.16&0.32 &0.21  &20 & 0.28&0.56 & 0.34 &20 &  0.30&0.60 &0.40 &20 & 0.28&0.56 & 0.38 &20   \\
  &SmoothGrad   & 0.12 &0.25 & 0.16 &20 &  0.24&0.48& 0.32 &20& \textbf{0.50}&\textbf{1.00} & \textbf{0.67} &20 & 0.22&0.43 & 0.29 &20   \\
  & IG    & 0.16&0.32 &0.22 &20 & 0.24&0.49 &0.33 &20 &  \textbf{0.50}&\textbf{1.00} &\textbf{0.67} &20 & 0.28&0.55 &0.37 &20   \\
\bottomrule
\end{tabular}
    }
    \label{tab:node_correctness_cora}
\end{table*}

\subsection{Plausibility}
Table~\ref{tab:graph_plaus_auprc} shows the \textit{Plausibility} scores computed for explanations of different GNN models. Recall that we compare explanations with human rationales to compute plausibility. The average size of human rationales over the test dataset is 165. To compute token level F1 score, we use mean as a threshold to generate hard masks from soft masks.

 We observe that all explainers assign the best plausibility scores to \gcn. \gin obtains the second-best plausibility scores. We also observe that the overall difference in the plausibility scores over models is relatively small, with some exceptions like the combination of \gin and \gnnexp. The corresponding explanation also has the largest size. This further highlights the issues of soft-hard mask conversion. AUPRC scores which directly use the soft masks are more stable.
 
 One surprising fact in these results is that even though other GNN models achieve higher test accuracy than \gin (see Table~\ref{tab:dataset stat for graph classification} in the Appendix). Overall, their explanations have similar plausibility as for \gin except for \gnnexp. In such cases, an application user might want to look in more detail at specific correctly labeled instances to check if the model imitates human reasoning.

\begin{table*}[htbp]
    \centering
    \caption{Plausibility for movie review dataset measured by auprc and F1 score (macro). |$\mathcal{S}$| represents the average size of the explanations generated by the explainers.}
    {\small
    \renewcommand*{\arraystretch}{1.0}
    \setlength{\tabcolsep}{3.5pt}
    \begin{tabular}{ll ccc c ccc c ccc c ccc}
\toprule
\multirow{2}{*}{Mask} & \multirow{2}{*}{Methods}   
 &     \multicolumn{3}{c}{GCN} & & \multicolumn{3}{c}{GAT} & & \multicolumn{3}{c}{GIN}& & \multicolumn{3}{c}{APPNP}   \\
& &auprc & F1 &|$\mathcal{S}$|&&auprc & F1 &|$\mathcal{S}$|  &&auprc &F1 &|$\mathcal{S}$| &&auprc &F1 &|$\mathcal{S}$| \\
\midrule
 Hard & PGM   & ---  & 0.42 & 25&& --- &   \textbf{0.43} & 25&& --- & \textbf{0.43} &25 && ---& \textbf{0.43} &25 \\
\midrule
\multirow{6}{*}{Soft} & \gnnexp &\underline{0.46}  & \textbf{0.54} &168 && 0.43 &\textbf{0.54}&149 && 0.45 &0.35 &410&& 0.45 & 0.53 &158  \\
 & Grad  & \underline{0.44} & \textbf{0.52} & 265&& 0.38 & 0.51&158 && 0.40 &\textbf{0.52} &156 && 0.38 & 0.50&255 \\
 & GradInput & \underline{0.39} &\textbf{0.51}& 221 && 0.37 & 0.50&154 && \underline{0.39} &\textbf{0.51} &154 && 0.37 & 0.50 &227 \\
  & SmoothGrad  & \underline{0.40} & \textbf{0.52} &219 && 0.37 &0.50&154 && \underline{0.40} & \textbf{0.52} &172 && 0.38& 0.50 &221  \\
 & IG  & 0.37 & 0.49&225 &&0.37 & 0.50 &188 && \underline{0.39} & \textbf{0.51} &186 && 0.38 & 0.50&219   \\
 & CAM & \underline{0.54}   &\textbf{0.61}&224 && 0.40 & 0.51&177&& 0.44    &0.55&156 && 0.44 & 0.53&195     \\
 & GradCAM & \underline{0.67}  &0.34&175 && \underline{0.67} &\textbf{0.35} &191 && \underline{0.67} & 0.34&166 && \underline{0.67} &0.34&188   \\

\bottomrule
\end{tabular}
    }
    \label{tab:graph_plaus_auprc}
\end{table*}

\section{Conclusion}
We develop a unified, modular, extendable benchmark called \ourbench to evaluate GNN explanations on four diverse axes: 1) \textit{faithfulness}, 2) \textit{sparsity}, 3) \textit{correctness}, and 4) \textit{plausibility}. 
We summarize the key insights of our experiments on \ourbench{} here:

\begin{itemize}[leftmargin=*]
\item Faithfulness measured via \textit{RDT-Fidelity} can be employed for a wide set of tasks and datasets. The only exception is when the input features are themselves non-interpretable. In that case, one should prefer structure-based explanations.
\item High RDT-Fidelity also implies high explanation stability.
    \item The  \textit{comprehensiveness} and \textit{sufficiency} measures should be used to evaluate the faithfulness of structure-based explanations where perturbing features might not be feasible.
    \item It is important to measure the sparsity of the explanation to avoid the extreme case of using the whole input as an explanation.
    \item  Correctness should be used with care, as injecting appropriate correlations to change a model's decision is not always straightforward. Approaches based on adversarial attacks to manipulate the GNN's prediction \cite{jin2021adversarial} can be adopted. 
\item Plausibility measures the joint utility of the explanation method and the trained GNN model with respect to human rationales.
This means that the loss of plausibility can be either due to human-incongruent correlations or the explainer, and it is impossible to disentangle them.     
\end{itemize}

\bibliographystyle{abbrv}

\clearpage
\appendix
\section*{Appendix}

\section{More details on datasets}
\label{sec:Tasks, Datasets and Baselines}

In this section we describe in detail the tasks and datasets used in our framework.

\subsection{Datasets}
\label{sec:datasets}
For \textit{node classification} task, we use citation datasets ~\cite{sen2008collective}, namely, \textsc{Cora}, \textsc{CiteSeer} and \textsc{PubMed}. In these citation datasets, each paper represents node and there is an edge between two papers if one cites other. Each node has its input features vectors and a label.
We use the semi-supervised setting for data spilt where only 20 nodes per label are used in training and 500 nodes for validation and 1000 nodes for test data. The details are available in Table~\ref{tab:dataset stat for node classification}. For correctness experiment, we use the fully supervised split where all the nodes except validation and test data are in training data.  

For \textit{graph classification} task, we use molecules datasets like PROTEINS~\cite{borgwardt2005protein} and MUTAG~\cite{debnath1991structure} and text dataset like Movie Reviews~\cite{zaidan2008modeling}. In PROTEINS dataset, each graph represents protein and task is to classify the protein into enzymes or non-enzymes. The MUTAG dataset contains 188 chemical compounds and task is to classify whether the compound has a mutagenic effect on a bacterium. The \movie is a text dataset which we transform into a graph dataset.  In the transformed  dataset each graph represents a movie review and the task is to classify the sentiment of the review as \textit{Positive or Negative}. The data statistics and the model accuracy are reported in Table~\ref{tab:dataset stat for graph classification}. We now describe the construction of \movie dataset. 

\begin{figure}[h!]
    \centering
    \includegraphics[width=12cm]{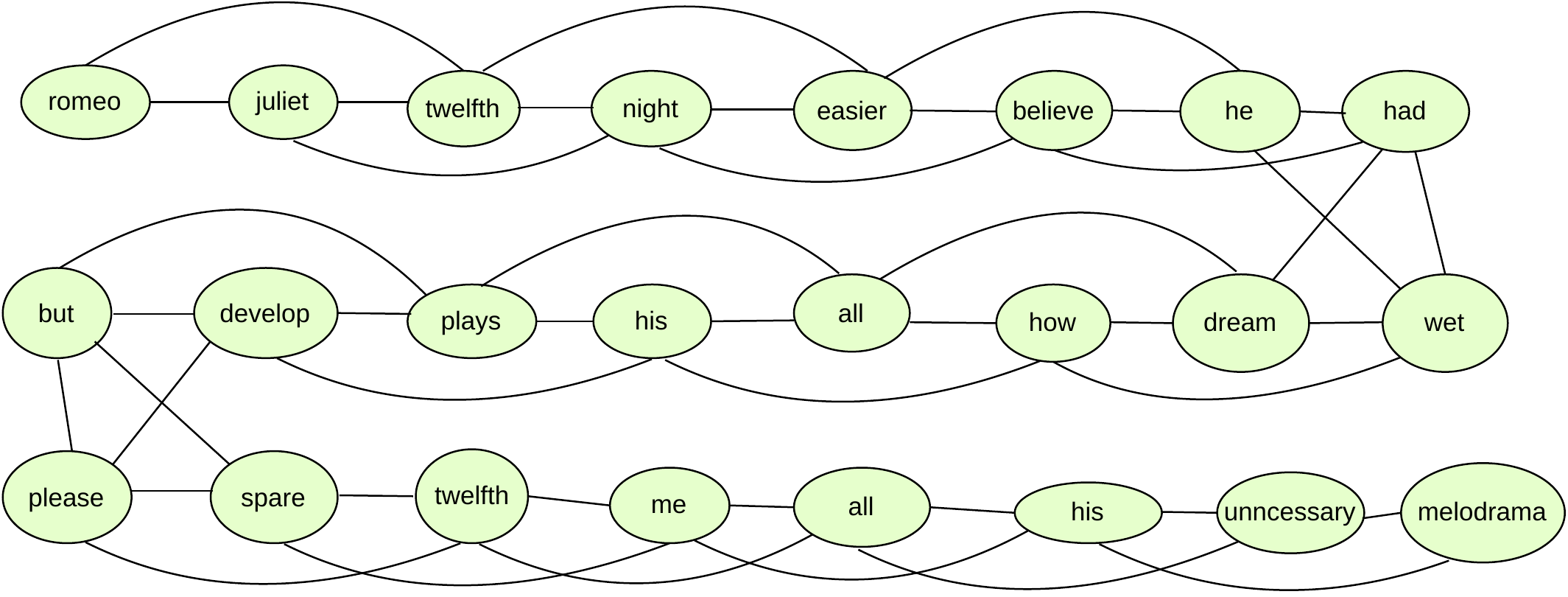} 
    \caption{An example for text to graph generation. The graph is corresponding to input sentences \textit{"? romeo and juliet ' , and ? the twelfth night ' . it is easier for me to believe that he had a wet dream and that 's how all his plays develop , but please spare me all of this unnecessary melodrama."}}
    \label{fig:text2graph}
\end{figure}

\mpara{Construction of Movie Reviews Dataset}. Each input instance or review is a passage of text, typically with multiple sentences. Each input review is annotated by humans that reflects the actual "human" reasons for predicting the sentiment of the review. These annotations are are extractive pieces of text and we call them human rationales. We transform sentences into graphs using the graph-of-words approach~\cite{rousseau2013graph}.
As a pre-processing step, we remove stopwords, such as "the" or "a". The complete list of used stopwords is included in our repository. Each word is represented as a node and all words within a sliding window of three are connected via edges. As features, we use the output of a pre-trained Glove model~\cite{pennington2014glove}. We then train a 2-layer GNN model with global average/attention pooling on the defined training set and record in Table~\ref{tab:dataset stat for graph classification} the achieved performance. Further details about dataset statistics and GNNs performances are available in Section~\ref{sec:Tasks, Datasets and Baselines}. Figure~\ref{fig:text2graph} provides an example of a graph from \movie dataset.

\subsection{Dataset statistics and Model Performance}

\begin{table*}[htbp]
    \centering
    \caption{Datasets statistics and model performance for node classification.}
    {\small
    \setlength{\tabcolsep}{3.5pt}
\begin{tabular}{lrrrrrrrrr}
\toprule
& & & & & \multicolumn{4}{c}{ \textbf{Test Accuracy} } \\
\cmidrule{6-9}
Dataset & Class & $d$ & $|V|$ & $|E|$ & \textbf{GCN} & \textbf{GAT} & \textbf{GIN} & \textbf{APPNP} \\
\midrule 
\textsc{Cora} & 7 & 1433 & 2708 & 10556 & $0.794$ & $0.791$ & $0.679$ & $0.799$ \\
\textsc{CiteSeer} & 6 & 3703 & 3327 & 9104 & $0.675$ & $0.673$ & $0.480$ & $0.663$ \\
\textsc{PubMed} & 3 & 500 & 19717 & 88648 & $0.782$ & $0.765$ & $0.590$ & $0.782$ \\
\bottomrule
\end{tabular}
    }
    \label{tab:dataset stat for node classification}
\end{table*}

\begin{table*}[htbp]
    \centering
    \caption{Datasets statistics and model performance for Graph classification.}
    {\small
    \setlength{\tabcolsep}{3.5pt}
\begin{tabular}{lrrrrrrrrr}
\toprule
& & & & & & \multicolumn{4}{c}{ \textbf{Test Accuracy} } \\
\cmidrule{7-10}
Dataset & \#Graphs & Class & $d$ & Avg. $|V|$ & Avg. $|E|$ & \textbf{GCN} & \textbf{GAT} & \textbf{GIN} & \textbf{APPNP} \\
\midrule 
MUTAG &188  & 2&7 & 17.9 & 39.6 & 0.76 & 0.76 &0.79 & 0.76 \\
PROTEINS &1113 &2 & 3 & 39.1 & 145.6 & 0.66 & 0.58 & 0.69 & 0.58 \\
Movie Reviews &2000& 2 & 300 & 500.8 &1997.5 &0.85 &0.85 & 0.78 & 0.82 \\
\bottomrule
\end{tabular}
    }
    \label{tab:dataset stat for graph classification}
\end{table*}

\section{Details of explainers}
\label{sec:Summary of the major Approaches}

In the following section, we  summarize the GNN explanation approaches currently implemented in \ourbench.

\subsection{Grad}
For a given node $v$ and the trained GNN model $f(\mathbf{x}_v, \mathcal{G},\phi)$, where $\phi$ is set of parameters , the gradient based approach, Grad\cite{imageclasssimonyan2013deep}, generate an explanation by assigning importance scores to the input features. The score corresponding to a feature reveals the importance of this feature in making prediction. The high score corresponds to high importance. The explanation for node $v$ is given by
\begin{equation}
  \mathcal{S}(\mathbf{x}_v)=  \frac{\partial f}{\partial \mathbf{x}_v}
\end{equation}
the gradients of $f$ with respect to input features $\mathbf{x}_v$.

\subsection{GradInput}
GradInput explanations are transformed version of Gradient based explanation by doing element-wise multiplication with input features. The explanation for node $v$ is given by:
\begin{equation}
  \mathcal{S}(\mathbf{x}_v)=  \frac{\partial f}{\partial \mathbf{x}_v}\odot{\mathbf{x}_{v}}
\end{equation}

\subsection{Integrated Gradient (IG)} The gradient based explanations are often noisy and can sometimes be insensitive with respect to input, for example if the learnt function $f$ is a straight line then the gradients are same with respect to different inputs. IG~\cite{sundararajan2017:integratedgrad} proposed the interpolation based gradient method which cumulates the gradients along straight path between input features vector $\mathbf{x}_v$ and a baseline vector $\mathbf{x}^\prime$ of all zeros or ones. The explanation for node $v$ is given by:
\begin{equation}
\mathcal{S}(\mathbf{x}_v) =\left(\mathbf{x}_v-\mathbf{x}^{\prime}\right) \times \int_{\alpha=0}^{1} \frac{\partial f\left(\mathbf{x}^{\prime}+\alpha \times\left(\mathbf{x}_v-\mathbf{x}^{\prime}\right)\right)}{\partial \mathbf{x}_v} d \alpha
\end{equation}

\subsection{SmoothGrad}
The SmoothGrad~\cite{smilkov2017smoothgrad} claims that the gradients of model $f$ may saturate, which means the importance score of a feature cause a strong effect globally but it shows very low effect locally. Also, the gradient based explanations are noisy. It proposed the mechanism of smoothing the noisy gradients by adding noise to the input. This process generate extra samples for training and the explanations are more robust to the noise. The explanations of SmoothGrad are given by:

\begin{equation}
\hat{\mathcal{S}}(\mathbf{x}_{v})=\frac{1}{n} \sum_{1}^{n} \mathcal{S}\left(\mathbf{x}_{v}+\mathcal{N}\left(0, \sigma^{2}\right)\right)
\end{equation}
where $n$ is the number of samples and $\mathcal{N}\left(0, \sigma^{2}\right)$ is the noise distribution. 

\subsection{CAM}
CAM~\cite{pope2019explainability} explains the graph classification tasks only. For graph classification tasks, we use an additional global average pooling (GAP) layer to generate the representation of the graph. For a graph $\mathcal{G}$, let $f_\mathcal{G}$ be the trained model for graph classification task which can be represented as:
\begin{equation}
    f_{\mathcal{G}}(\mathcal{A}, \mathcal{X}) = \sigma(\operatorname{GAP}(\mathbf{x}^{(L)}_{u} \mid u \in \mathcal{G}), \textit{w})
\end{equation}
where $L$ represents the final layer of GNN, $\mathbf{x}_{u}^{(L)}$ is the representation of node $u$ at layer $L$ and $\textit{w}$ represents the trainable parameters for the classification layer and $\sigma$ is the activation function. CAM assumes that the final layer representations encode the input feature behavior corresponds to prediction and the input feature importance is given by weighted sum of different features. The explanation of the graph $\mathcal{G}$ is given by:
\begin{equation}
\mathcal{S(G)}=\operatorname{ReLU}\left(\sum_{k} w_{k} \mathcal{X}_{k}^{(L)}\right)
\end{equation}
where $\mathcal{X}_{k}^{(L)}$ represents the $k^{th}$ feature at layer $L$. 

\subsection{GradCAM}
GradCAM~\cite{pope2019explainability} is an extended version of CAM designed for graph classification based GNN models which does not need the global average pooling (GAP) layer. It also uses the final representation to generate the importance score. It assigns the gradient based weights to the representation.  The weight for $k^{th}$ feature at layer $l$ is computed as:
\begin{equation}
\alpha_{k}^{(l)}=\frac{1}{N} \sum_{n=1}^{N} \frac{\partial f_\mathcal{G}}{\partial \mathcal{X}_{k}^{(l)}}
\end{equation}
where $N$ is the number of nodes in graph $\mathcal{G}$. Finally the explanation of the graph is given by:
\begin{equation}
\mathcal{S(G)}=\operatorname{ReLU}\left(\sum_{k} \alpha_{k}^{l} \mathcal{X}_{k}^{(l)})\right)
\end{equation}

\subsection{Zorro}
Zorro\cite{funke2021zorro} is a greedy-combinatorial algorithm, which optimizes \textit{RDT-Fidelity}, to receive valid, sparse, and stable explanations. Zorro iteratively extends the explanation with the feature or node, which yields the highest increase in \textit{RDT-Fidelity}. The addition of features and nodes is stopped, when a pre-defined threshold is reached. Hence, it automatically determines the explanation size. For the formulation and details of \textit{RDT-Fidelity} see Section~\ref{subsubsec:Faithfulness}. The authors in~\cite{funke2021zorro} evaluated two different \textit{RDT-Fidelity} thresholds. Here, we performed the experiments using the higher threshold of $\tau=0.98$. Zorro is designed for node classification and is model agnostic. The explanations consists of hard feature and hard node masks. 

\subsection{PGM}
PGM-Explainer (\pgm)~\cite{vu2020pgm} is a surrogate based method which fits a simple and interpretable model on a sampled local dataset. The first step is data generation which is collection of perturbed input for a given target node and its prediction. Secondly the variable selector removes the irrelevant samples from the generated data and finally a Bayesian network is trained on the sampled data and the explanations generated by Bayesian network are treated as explanations of GNN model for the target node. \pgm retrieves hard masks over the nodes. \pgm explains both node and graph classification tasks.   

\subsection{\gnnexp} 
For a given graph $\mathcal{G}$, \gnnexp~\cite{ying2019:gnnexplainer} learns subgraph $\mathcal{G^\prime}\subseteq \mathcal{G}$ which contains the important graph structure (mainly edges) and import features which are responsible for the prediction. For the subgraph, \gnnexp learns the soft masks over edges and features which are optimized using the mutual information between the prediction of GNN on $\mathcal{G}^{\prime}$ and the prediction ($Y$) of GNN on $\mathcal{G}$. Mathematically, 

\begin{equation}
\max _{\mathcal{G}^{\prime}} M I\left(Y,\mathcal{G}^{\prime}\right) = 
H(Y)-H\left(Y \mid \mathcal{G}=\mathcal{G}^{\prime}\right)
\end{equation}
where $H(.)$ is the entropy. \gnnexp learns masks locally and applicable to all graph based tasks like graph classification, node classification and link prediction.

\section{Results of RDT-Fidelity on Graph classification}
In Table~\ref{tab:graph_fidelity}, we report \textit{RDT-Fidelity} for graph classification task.

\begin{table*}[h!]
    \centering
     \caption{\textit{RDT-Fidelity} for graph classification.}

    {\small
    \renewcommand*{\arraystretch}{1.0}
    \setlength{\tabcolsep}{2.9pt}
    \begin{tabular}{ll ccc ccc ccc }
\toprule

\multirow{2}{*}{Mask} & \multirow{2}{*}{Methods} & \multicolumn{4}{c}{MUTAG} & \multicolumn{4}{c}{PROTEINS}  \\
 &  &  \small GCN &\small   GAT & \small  GIN & \small APPNP & \small GCN & \small  GAT & \small  GIN & \small APPNP  \\
\midrule
 \footnotesize
\multirow{1}{*}{Hard}& PGM      &  0.58 & \textbf{1.00} &  0.60 & \textbf{1.00} &     \textbf{0.97} & \textbf{1.00} & 0.89 & \textbf{1.00}  \\
\midrule
\multirow{7}{*}{Soft}& GNNExp   &  \textbf{0.73} &  0.96 & \textbf{0.75} &  0.99 & 0.38 &0.46 & 0.23 &  \textbf{1.00}\\
 & Grad           & 0.39 & 0.86 &  0.58 &  0.96 &  0.93 &  0.90 &  \textbf{0.90} & 0.94  \\
 & GradInput     &  0.38 & 0.87 &  0.58 &  0.96 &   0.93 &  0.91 &  \textbf{0.90} &  0.94 \\
  & SmoothGrad      &  0.38 &  0.86 &  0.59 &  0.96 &   0.93 &  0.90 &  \textbf{0.90} &  0.94  \\
 & IG      &  0.38 &  0.85 &  0.59 & 0.96 & 0.93 &  0.92 &  \textbf{0.90} &  0.94  \\
 & CAM      &  0.39 &  0.85 &  0.59 &  0.97 &  0.93&  0.91 &  \textbf{0.90} &  0.95  \\
 & GradCAM     &  0.38 &  0.85 &  0.58 & 0.96 & 0.93 &  0.90 &  0.89 &  0.95  \\

\bottomrule
\end{tabular}
    }
    \label{tab:graph_fidelity}
\end{table*}

\section{Results of Correctness on \citeseer}
We report correctness for \citeseer dataset in Tablw~\ref{tab:node_correctness_citeseer}.

\begin{table*}[htbp]
    \centering
    \caption{Correctness  of the explanation for node classification on \textsc{CiteSeer} dataset. We use $k=20$.}
    {\small
    \setlength{\tabcolsep}{2.5pt}
    \begin{tabular}{ll cccc cccc cccc cccc}
\toprule
\multirow{3}{*}{Mask}&\multirow{3}{*}{Methods} &  \multicolumn{16}{c}{\textsc{CiteSeer}} \\
&  &\multicolumn{4}{c}{GCN} & \multicolumn{4}{c}{GAT}  &\multicolumn{4}{c}{GIN} & \multicolumn{4}{c}{APPNP} \\
&  & P@k & R@k &F1 & |$\mathcal{S}$| & P@k & R@k &F1 & |$\mathcal{S}$| & P@k & R@k &F1 & |$\mathcal{S}$| & P@k & R@k &F1 & |$\mathcal{S}$|\\
\midrule
\multirow{2}{*}{Hard} & ZORRO    & 0.17 &0.59& 0.24 &  36 & 0.31&0.41& 0.33 &  14 &  0.19 & 0.53 & 0.25 &  35& 0.20 &0.61& 0.25 &  40  \\
  & PGM     &  0.17&0.34 &  0.23 & 20 & 0.18&0.36 & 0.24 &  20 &  0.19 &0.39 &0.25 &  20 &  0.17&0.34 &  0.23 &  20 \\
 \midrule 
\multirow{5}{*}{Soft} & \gnnexp   & \textbf{0.47}&\textbf{0.94} &  0.63 &  20 & \textbf{0.46}&\textbf{0.92} &  0.62 &  20 & \textbf{0.47}&\textbf{0.95} &  0.63 & 20 &  \textbf{0.46}&\textbf{0.92}&  0.61 &  20 \\
 & Grad   &  0.04&0.08 &  0.06 &  20& 0.28&0.56  &  0.38 &  20 & 0.23&0.46 &  0.31 & 20 & 0.23&0.46&  0.31 &  20 \\
 & GradInput  &0.02&0.04 &  0.02 &  20 & 0.25&0.50 &  0.29 &  20& 0.26&0.51 & 0.34 &  20&  0.09&0.17 &  0.12 &  20  \\
  & IG    & 0.02&0.04& 0.03 &  20 & 0.23&0.46 &  0.31 &  20 & 0.37&0.73 & 0.49 &  20&  0.13&0.25&  0.17 &  20  \\
 &SmoothGrad   & 0.01&0.02& 0.02 &  20 & 0.32&0.64&  0.43 & 20 &  0.29&0.58 & 0.38 & 20 &  0.06&0.12 &  0.08 &  20  \\

\bottomrule
\end{tabular}
    }
    \label{tab:node_correctness_citeseer}
\end{table*}
\section{Different strategies to compute hard mask from soft mask to measure Correctness}
\label{sec: Why evaluation with soft masks is difficult?}

We report correctness for \cora and \citeseer in Table~\ref{tab:node_correctness_cora_mean} and~\ref{tab:node_correctness_citeseer_mean} with mean as threshold for generating hard masks.

\begin{table*}[htbp]
    \centering
    \caption{Correctness  of the explanation on \textsc{Cora} dataset. We use mean as threshold to generate hard masks. |$\mathcal{S}$| represents size of the explanations.}
    {\small
    \renewcommand*{\arraystretch}{0.8}
    \setlength{\tabcolsep}{4.5pt}
    \begin{tabular}{ll ccc ccc ccc ccc}
\toprule
\multirow{3}{*}{Mask}&\multirow{3}{*}{Methods} &  \multicolumn{12}{c}{\textsc{Cora}} \\
&  &\multicolumn{3}{c}{GCN} & \multicolumn{3}{c}{GAT} &\multicolumn{3}{c}{GIN} & \multicolumn{3}{c}{APPNP} \\
& & Pre. & Rec.&|$\mathcal{S}$| & Pre. & Rec.&|$\mathcal{S}$| & Pre. & Rec.&|$\mathcal{S}$| & Pre.& Rec.& |$\mathcal{S}$|\\
\midrule
\multirow{2}{*}{Hard} & ZORRO    &0.19 & 0.80 & 45 & 0.25 &0.83  & 40 & 0.26  &0.45 & 33 &  0.22 & 0.79 & 38   \\
  & PGM      & 0.11&0.22& 20& 0.18&0.36&20& 0.18&0.36 &20& 0.19&0.38&20   \\
\midrule
\multirow{5}{*}{Soft} & GNNExplainer      & 0.11  &  1.00 & 108&  0.11 & 1.00 &109&0.26 & 1.00&39 &0.11 &  1.00 &107  \\
 & Grad     & 0.12 &  0.99 &94& 0.12  &  1.00&91 &0.24 & 0.90&38 &  0.12 &  0.99&94 \\
 & GradInput   & 0.11  &  1.00&98 &  0.11 & 1.00&106 &0.26 &1.00&39 & 0.11 &  1.00 &99\\
  & IG       & 0.11  & 1.00 &98&  0.11 &  1.00&100 &0.26 & 1.00 &39& 0.11 &  1.00&101 \\
 &SmoothGrad     & 0.11  & 1.00&104 & 0.11 & 1.00 &102.&0.26 & 1.00 &39&  0.11 & 1.00&102 \\
\bottomrule
\end{tabular}
    }
    \label{tab:node_correctness_cora_mean}
\end{table*}

\begin{table*}[htbp]
    \centering
    \caption{Correctness  of the explanation for \textsc{CiteSeer} dataset. We use mean as threshold to generate hard masks. |$\mathcal{S}$| represents size of the explanations.}
    {\small
    \renewcommand*{\arraystretch}{0.8}
    \setlength{\tabcolsep}{4.5pt}
    \begin{tabular}{ll ccc ccc ccc ccc}
\toprule
\multirow{3}{*}{Mask}&\multirow{3}{*}{Methods} &  \multicolumn{12}{c}{\textsc{CiteSeer}} \\
&  &\multicolumn{3}{c}{GCN} & \multicolumn{3}{c}{GAT}  &\multicolumn{3}{c}{GIN} & \multicolumn{3}{c}{APPNP} \\
& & Pre. & Rec.&|$\mathcal{S}$| & Pre. & Rec.&|$\mathcal{S}$| & Pre. & Rec.&|$\mathcal{S}$| & Pre.& Rec. &|$\mathcal{S}$|\\
\midrule
\multirow{2}{*}{Hard} & ZORRO    & 0.17 &0.59 & 36 & 0.31&0.41&14 &  0.19&0.53&35 & 0.20&0.61&40 \\
 & PGM     &  0.17&0.34 &20&0.18&0.36  &20&  0.17&0.34  &20&  0.17&0.34  &20\\
\midrule
\multirow{5}{*}{Soft} & GNNExplainer     & 0.11 &  1.00 &99& 0.10 &  1.00&111 &0.19 &1.00&60 & 0.10 & 1.00&106  \\
 & Grad      & 0.11  &  0.97&92 &  0.11 & 0.99&99 &0.19 & 0.98&58 & 0.11 &  0.99&97\\
 & GradInput    & 0.11  & 1.00&95 &  0.10 & 1.00&106 &0.19 & 1.00&59 & 0.11 & 1.00&102 \\
  & IG      & 0.11  &  1.00&95 &  0.10 &  1.00&106 &0.19 &1.00&59 & 0.11 & 1.00&101 \\
 &SmoothGrad     & 0.11 & 1.00 &99&  0.10 & 1.00&110 &0.19&1.00&60 &  0.10 & 1.00&106\\

\bottomrule
\end{tabular}
    }
    \label{tab:node_correctness_citeseer_mean}
\end{table*}

\begin{figure}[]
        \begin{subfigure}{0.50\columnwidth}
            \begin{tikzpicture}
		\begin{axis}[
			width  = \textwidth,
			height = 0.8\textwidth,
			major x tick style = transparent,
			grid = major,
		    grid style = {dashed, gray!20},
			ylabel = {ma-F1},
			title={GCN},
			symbolic x coords={5, 10, 15, 20, 25},
			xtick = data,
			enlarge x limits=0.25,
			]
			\addplot [color=black, mark=*, line width = 1pt] table [x index=0, y index=1, col sep = comma] {plots/movie_reviews/ma-f1_GCN.txt};\label{plot_GCN_gnne}
			
			\addplot [color=red, mark=square*, line width = 1pt] table [x index=0, y index=2, col sep = comma] {plots/movie_reviews/ma-f1_GCN.txt};\label{plot_GCN_grad}
			
			\addplot [color=green, mark=diamond*, line width = 1pt] table [x index=0, y index=3, col sep = comma] {plots/movie_reviews/ma-f1_GCN.txt};\label{plot_GCN_gradinput}
			
			\addplot [color=brown, mark=triangle*, line width = 1pt] table [x index=0, y index=4, col sep = comma] {plots/movie_reviews/ma-f1_GCN.txt};\label{plot_GCN_smoothgrad}
			
			\addplot [color=blue, mark=square, line width = 1pt] table [x index=0, y index=5, col sep = comma] {plots/movie_reviews/ma-f1_GCN.txt};\label{plot_GCN_ig}
			
			\addplot [color=cyan, mark=triangle, line width = 1pt] table [x index=0, y index=6, col sep = comma] {plots/movie_reviews/ma-f1_GCN.txt};\label{plot_GCN_cam}
			
			\addplot [color=magenta, mark=diamond, line width = 1pt] table [x index=0, y index=7, col sep = comma] {plots/movie_reviews/ma-f1_GCN.txt};\label{plot_GCN_gradcam}
			
		\end{axis}
	\end{tikzpicture}
        \end{subfigure}
	    \hfill
	    \begin{subfigure}{0.50\textwidth}
            \begin{tikzpicture}
		\begin{axis}[
			width  = \textwidth,
			height = 0.8\textwidth,
			major x tick style = transparent,
			grid = major,
		    grid style = {dashed, gray!20},
			ylabel = {mi-F1},
			title={GCN},
			symbolic x coords={5, 10, 15, 20, 25},
			xtick = data,
			enlarge x limits=0.25,
			]
			\addplot [color=black, mark=*, line width = 1pt] table [x index=0, y index=1, col sep = comma] {plots/movie_reviews/mi-f1_GCN.txt};\label{plot_GCN_gnne}
			
			\addplot [color=red, mark=square*, line width = 1pt] table [x index=0, y index=2, col sep = comma] {plots/movie_reviews/mi-f1_GCN.txt};\label{plot_GCN_grad}
			
			\addplot [color=green, mark=diamond*, line width = 1pt] table [x index=0, y index=3, col sep = comma] {plots/movie_reviews/mi-f1_GCN.txt};\label{plot_GCN_gradinput}
			
			\addplot [color=brown, mark=triangle*, line width = 1pt] table [x index=0, y index=4, col sep = comma] {plots/movie_reviews/mi-f1_GCN.txt};\label{plot_GCN_smoothgrad}
			
			\addplot [color=blue, mark=square, line width = 1pt] table [x index=0, y index=5, col sep = comma] {plots/movie_reviews/mi-f1_GCN.txt};\label{plot_GCN_ig}
			
			\addplot [color=cyan, mark=triangle, line width = 1pt] table [x index=0, y index=6, col sep = comma] {plots/movie_reviews/mi-f1_GCN.txt};\label{plot_GCN_cam}
			
			\addplot [color=magenta, mark=diamond, line width = 1pt] table [x index=0, y index=7, col sep = comma] {plots/movie_reviews/mi-f1_GCN.txt};\label{plot_GCN_gradcam}			
			
		\end{axis}
	\end{tikzpicture}
        \end{subfigure}
 \begin{subfigure}{0.50\columnwidth}
            \begin{tikzpicture}
		\begin{axis}[
			width  = \textwidth,
			height = 0.8\textwidth,
			major x tick style = transparent,
			grid = major,
		    grid style = {dashed, gray!20},
			ylabel = {ma-F1},
			title={GAT},
			symbolic x coords={5, 10, 15, 20, 25},
			xtick = data,
			enlarge x limits=0.25,
			]
			
			\addplot [color=black, mark=*, line width = 1pt] table [x index=0, y index=1, col sep = comma] {plots/movie_reviews/ma-f1_GAT.txt};\label{plot_GAT_gnne}
			
			\addplot [color=red, mark=square*, line width = 1pt] table [x index=0, y index=2, col sep = comma] {plots/movie_reviews/ma-f1_GAT.txt};\label{plot_GAT_grad}
			
			\addplot [color=green, mark=diamond*, line width = 1pt] table [x index=0, y index=3, col sep = comma] {plots/movie_reviews/ma-f1_GAT.txt};\label{plot_GAT_gradinput}
			
			\addplot [color=brown, mark=triangle*, line width = 1pt] table [x index=0, y index=4, col sep = comma] {plots/movie_reviews/ma-f1_GAT.txt};\label{plot_GAT_smoothgrad}
			
			\addplot [color=blue, mark=square, line width = 1pt] table [x index=0, y index=5, col sep = comma] {plots/movie_reviews/ma-f1_GAT.txt};\label{plot_GAT_ig}
			
			\addplot [color=cyan, mark=triangle, line width = 1pt] table [x index=0, y index=6, col sep = comma] {plots/movie_reviews/ma-f1_GAT.txt};\label{plot_GAT_cam}
			
			\addplot [color=magenta, mark=diamond, line width = 1pt] table [x index=0, y index=7, col sep = comma] {plots/movie_reviews/ma-f1_GAT.txt};\label{plot_GAT_gradcam}
			
		\end{axis}
	\end{tikzpicture}
        \end{subfigure}    
\begin{subfigure}{0.50\textwidth}
            \begin{tikzpicture}
		\begin{axis}[
			width  = \textwidth,
			height = 0.8\textwidth,
			major x tick style = transparent,
			grid = major,
		    grid style = {dashed, gray!20},
			ylabel = {mi-F1},
			title={GAT},
			symbolic x coords={5, 10, 15, 20, 25},
			xtick = data,
			enlarge x limits=0.25,
			]
			
			\addplot [color=black, mark=*, line width = 1pt] table [x index=0, y index=1, col sep = comma] {plots/movie_reviews/mi-f1_GAT.txt};\label{plot_GAT_gnne}
			
			\addplot [color=red, mark=square*, line width = 1pt] table [x index=0, y index=2, col sep = comma] {plots/movie_reviews/mi-f1_GAT.txt};\label{plot_GAT_grad}
			
			\addplot [color=green, mark=diamond*, line width = 1pt] table [x index=0, y index=3, col sep = comma] {plots/movie_reviews/mi-f1_GAT.txt};\label{plot_GAT_gradinput}
			
			\addplot [color=brown, mark=triangle*, line width = 1pt] table [x index=0, y index=4, col sep = comma] {plots/movie_reviews/mi-f1_GAT.txt};\label{plot_GAT_smoothgrad}
			
			\addplot [color=blue, mark=square, line width = 1pt] table [x index=0, y index=5, col sep = comma] {plots/movie_reviews/mi-f1_GAT.txt};\label{plot_GAT_ig}
			
			\addplot [color=cyan, mark=triangle, line width = 1pt] table [x index=0, y index=6, col sep = comma] {plots/movie_reviews/mi-f1_GAT.txt};\label{plot_GAT_cam}
			
			\addplot [color=magenta, mark=diamond, line width = 1pt] table [x index=0, y index=7, col sep = comma] {plots/movie_reviews/mi-f1_GAT.txt};\label{plot_GAT_gradcam}
			
		\end{axis}
	\end{tikzpicture}
        \end{subfigure}      
\begin{subfigure}{0.50\textwidth}
            \begin{tikzpicture}
		\begin{axis}[
			width  = \textwidth,
			height = 0.8\textwidth,
			major x tick style = transparent,
			grid = major,
		    grid style = {dashed, gray!20},
			ylabel = {ma-F1},
			title={GIN},
			symbolic x coords={5, 10, 15, 20, 25},
			xtick = data,
			enlarge x limits=0.25,
			]
			
			\addplot [color=black, mark=*, line width = 1pt] table [x index=0, y index=1, col sep = comma] {plots/movie_reviews/ma-f1_GIN.txt};\label{plot_GIN_gnne}
			
			\addplot [color=red, mark=square*, line width = 1pt] table [x index=0, y index=2, col sep = comma] {plots/movie_reviews/ma-f1_GIN.txt};\label{plot_GIN_grad}
			
			\addplot [color=green, mark=diamond*, line width = 1pt] table [x index=0, y index=3, col sep = comma] {plots/movie_reviews/ma-f1_GIN.txt};\label{plot_GIN_gradinput}
			
			\addplot [color=brown, mark=triangle*, line width = 1pt] table [x index=0, y index=4, col sep = comma] {plots/movie_reviews/ma-f1_GIN.txt};\label{plot_GIN_smoothgrad}
			
			\addplot [color=blue, mark=square, line width = 1pt] table [x index=0, y index=5, col sep = comma] {plots/movie_reviews/ma-f1_GIN.txt};\label{plot_GIN_ig}
			
			\addplot [color=cyan, mark=triangle, line width = 1pt] table [x index=0, y index=6, col sep = comma] {plots/movie_reviews/ma-f1_GIN.txt};\label{plot_GIN_cam}
			
			\addplot [color=magenta, mark=diamond, line width = 1pt] table [x index=0, y index=7, col sep = comma] {plots/movie_reviews/ma-f1_GIN.txt};\label{plot_GIN_gradcam}
			
		\end{axis}
	\end{tikzpicture}
        \end{subfigure}         
\begin{subfigure}{0.50\textwidth}
            \begin{tikzpicture}
		\begin{axis}[
			width  = \textwidth,
			height = 0.8\textwidth,
			major x tick style = transparent,
			grid = major,
		    grid style = {dashed, gray!20},
			ylabel = {mi-F1},
			title={GIN},
			symbolic x coords={5, 10, 15, 20, 25},
			xtick = data,
			enlarge x limits=0.25,
			]
			
			\addplot [color=black, mark=*, line width = 1pt] table [x index=0, y index=1, col sep = comma] {plots/movie_reviews/mi-f1_GIN.txt};\label{plot_GIN_gnne}
			
			\addplot [color=red, mark=square*, line width = 1pt] table [x index=0, y index=2, col sep = comma] {plots/movie_reviews/mi-f1_GIN.txt};\label{plot_GIN_grad}
			
			\addplot [color=green, mark=diamond*, line width = 1pt] table [x index=0, y index=3, col sep = comma] {plots/movie_reviews/mi-f1_GIN.txt};\label{plot_GIN_gradinput}
			
			\addplot [color=brown, mark=triangle*, line width = 1pt] table [x index=0, y index=4, col sep = comma] {plots/movie_reviews/mi-f1_GIN.txt};\label{plot_GIN_smoothgrad}
			
			\addplot [color=blue, mark=square, line width = 1pt] table [x index=0, y index=5, col sep = comma] {plots/movie_reviews/mi-f1_GIN.txt};\label{plot_GIN_ig}
			
			\addplot [color=cyan, mark=triangle, line width = 1pt] table [x index=0, y index=6, col sep = comma] {plots/movie_reviews/mi-f1_GIN.txt};\label{plot_GIN_cam}
			
			\addplot [color=magenta, mark=diamond, line width = 1pt] table [x index=0, y index=7, col sep = comma] {plots/movie_reviews/mi-f1_GIN.txt};\label{plot_GIN_gradcam}
			
		\end{axis}
	\end{tikzpicture}
        \end{subfigure}         
\begin{subfigure}{0.50\textwidth}
            \begin{tikzpicture}
		\begin{axis}[
			width  = \textwidth,
			height = 0.8\textwidth,
			major x tick style = transparent,
			grid = major,
		    grid style = {dashed, gray!20},
			xlabel = {topk},
			ylabel = {ma-F1},
			title={APPNP},
			symbolic x coords={5, 10, 15, 20, 25},
			xtick = data,
			enlarge x limits=0.25,
			]
			
			\addplot [color=black, mark=*, line width = 1pt] table [x index=0, y index=1, col sep = comma] {plots/movie_reviews/ma-f1_APPNP.txt};\label{plot_APPNP_gnne}
			
			\addplot [color=red, mark=square*, line width = 1pt] table [x index=0, y index=2, col sep = comma] {plots/movie_reviews/ma-f1_APPNP.txt};\label{plot_APPNP_grad}
			
			\addplot [color=green, mark=diamond*, line width = 1pt] table [x index=0, y index=3, col sep = comma] {plots/movie_reviews/ma-f1_APPNP.txt};\label{plot_APPNP_gradinput}
			
			\addplot [color=brown, mark=triangle*, line width = 1pt] table [x index=0, y index=4, col sep = comma] {plots/movie_reviews/ma-f1_APPNP.txt};\label{plot_APPNP_smoothgrad}
			
			\addplot [color=blue, mark=square, line width = 1pt] table [x index=0, y index=5, col sep = comma] {plots/movie_reviews/ma-f1_APPNP.txt};\label{plot_APPNP_ig}
			
			\addplot [color=cyan, mark=triangle, line width = 1pt] table [x index=0, y index=6, col sep = comma] {plots/movie_reviews/ma-f1_APPNP.txt};\label{plot_APPNP_cam}
			
			\addplot [color=magenta, mark=diamond, line width = 1pt] table [x index=0, y index=7, col sep = comma] {plots/movie_reviews/ma-f1_APPNP.txt};\label{plot_APPNP_gradcam}
			
		\end{axis}
	\end{tikzpicture}
        \end{subfigure}         
\begin{subfigure}{0.50\textwidth}
            \begin{tikzpicture}
		\begin{axis}[
			width  = \textwidth,
			height = 0.8\textwidth,
			major x tick style = transparent,
			grid = major,
		    grid style = {dashed, gray!20},
			xlabel = {topk},
			ylabel = {mi-f1},
			title={APPNP},
			symbolic x coords={5, 10, 15, 20, 25},
			xtick = data,
			enlarge x limits=0.25,
			]
			
			\addplot [color=black, mark=*, line width = 1pt] table [x index=0, y index=1, col sep = comma] {plots/movie_reviews/mi-f1_APPNP.txt};\label{plot_APPNP_gnne}
			
			\addplot [color=red, mark=square*, line width = 1pt] table [x index=0, y index=2, col sep = comma] {plots/movie_reviews/mi-f1_APPNP.txt};\label{plot_APPNP_grad}
			
			\addplot [color=green, mark=diamond*, line width = 1pt] table [x index=0, y index=3, col sep = comma] {plots/movie_reviews/mi-f1_APPNP.txt};\label{plot_APPNP_gradinput}
			
			\addplot [color=brown, mark=triangle*, line width = 1pt] table [x index=0, y index=4, col sep = comma] {plots/movie_reviews/mi-f1_APPNP.txt};\label{plot_APPNP_smoothgrad}
			
			\addplot [color=blue, mark=square, line width = 1pt] table [x index=0, y index=5, col sep = comma] {plots/movie_reviews/mi-f1_APPNP.txt};\label{plot_APPNP_ig}
			
			\addplot [color=cyan, mark=triangle, line width = 1pt] table [x index=0, y index=6, col sep = comma] {plots/movie_reviews/mi-f1_APPNP.txt};\label{plot_APPNP_cam}
			
			\addplot [color=magenta, mark=diamond, line width = 1pt] table [x index=0, y index=7, col sep = comma] {plots/movie_reviews/mi-f1_APPNP.txt};\label{plot_APPNP_gradcam}
			
		\end{axis}
	\end{tikzpicture}
        \end{subfigure}         
\caption{Plausibility for \movie. GNNExp~\ref{plot_GCN_gnne}~~Grad~\ref{plot_APPNP_grad}~~GradInput~\ref{plot_APPNP_gradinput}~~SmoothGrad~\ref{plot_GCN_smoothgrad}~~IG~\ref{plot_GCN_ig}~~CAM~\ref{plot_APPNP_cam}~~GradCAM~\ref{plot_APPNP_gradcam}}
\label{fig:movie_review_topk}
\end{figure}
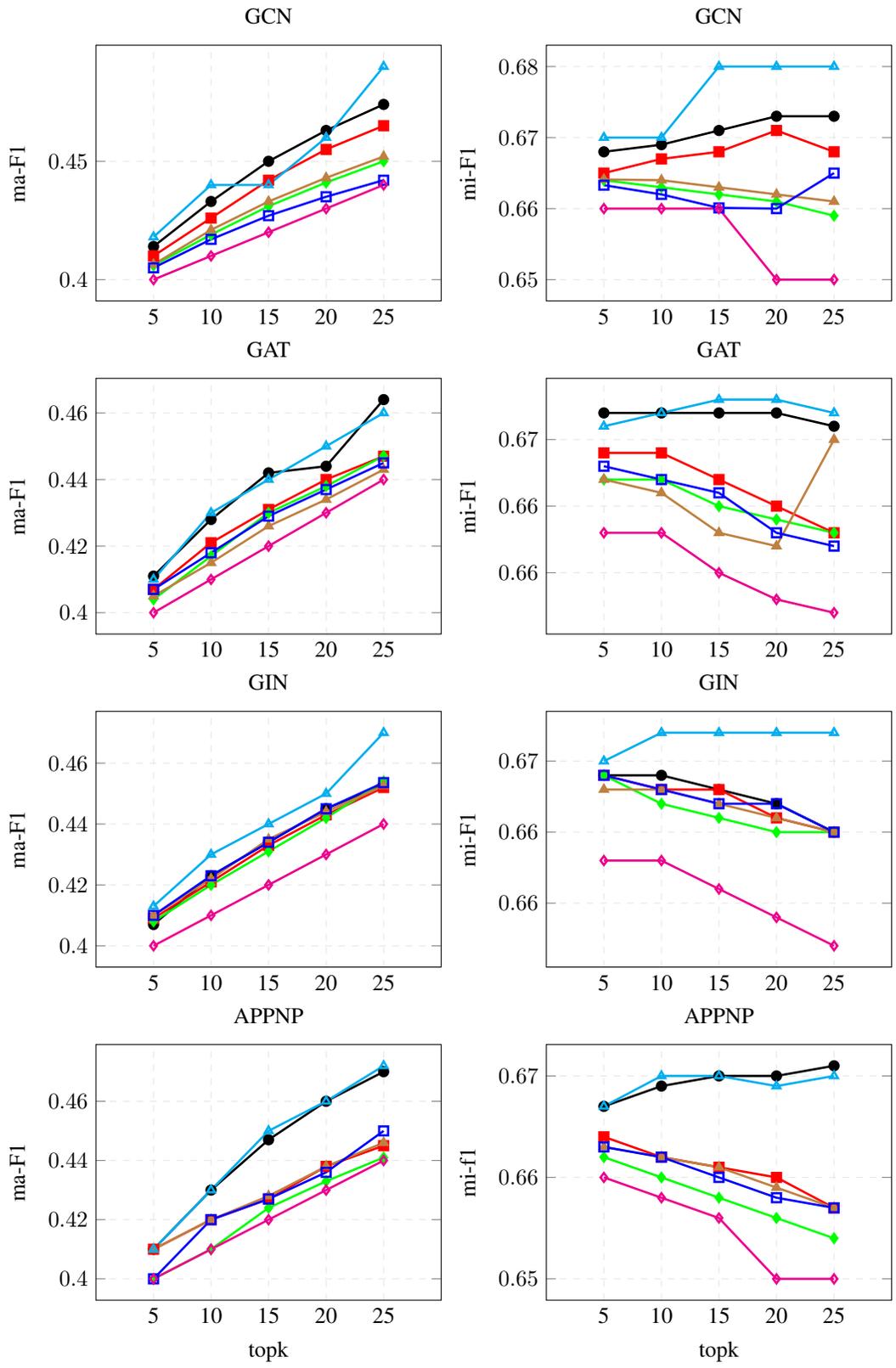

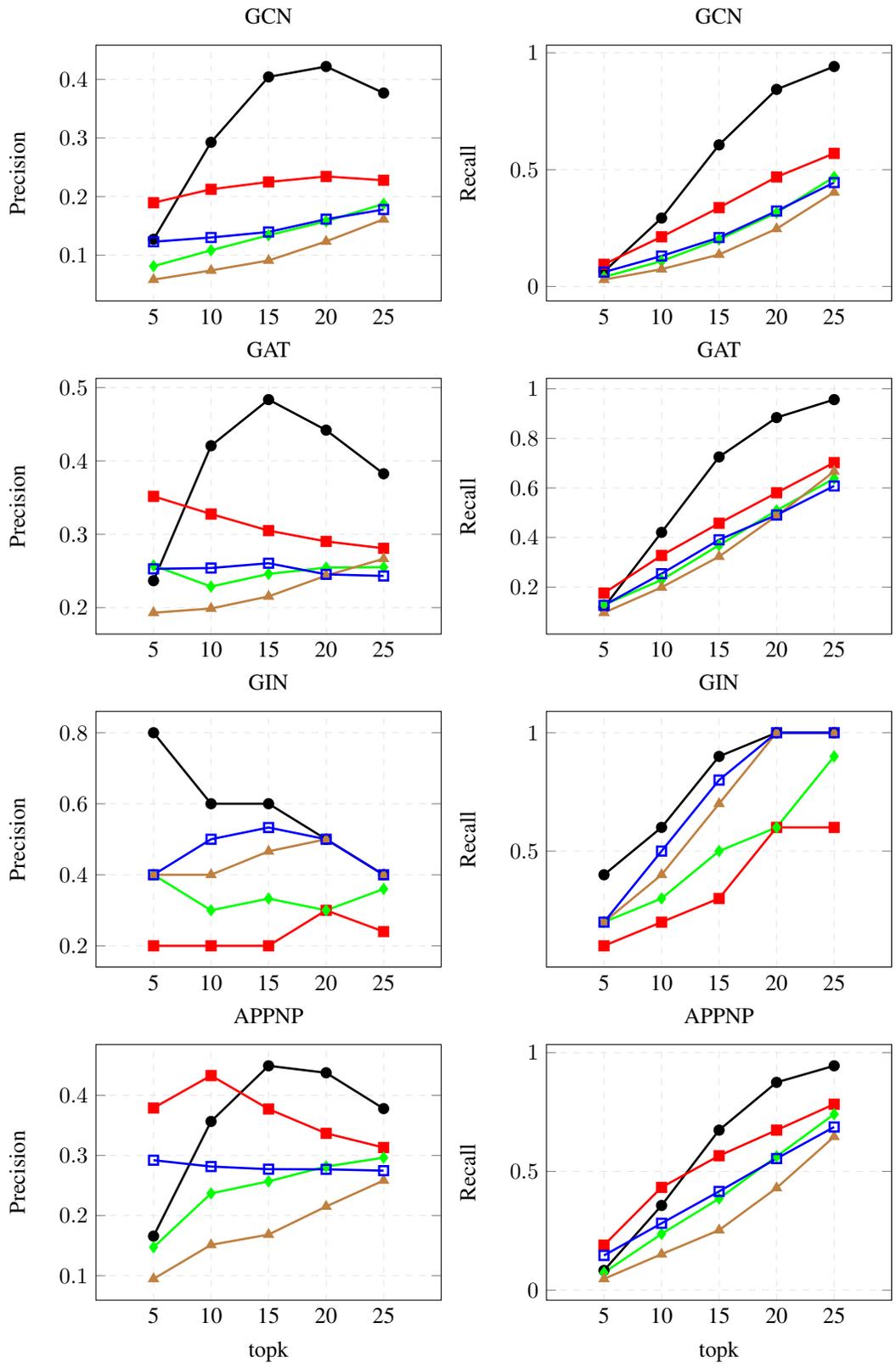
\begin{figure}[]
        \begin{subfigure}{0.5\textwidth}
            \begin{tikzpicture}
		\begin{axis}[
			width  = \textwidth,
			height = 0.8\textwidth,
			major x tick style = transparent,
			grid = major,
		    grid style = {dashed, gray!20},
			ylabel = {Precision},
			title={GCN},
			symbolic x coords={5, 10, 15, 20, 25},
			xtick = data,
			enlarge x limits=0.25,
			]
			\addplot [color=black, mark=*, line width = 1pt] table [x index=0, y index=1, col sep = comma] {plots/Correctness/precision_cora_GCN.txt};\label{pre_cora_GCN_gnne}
			
			\addplot [color=red, mark=square*, line width = 1pt] table [x index=0, y index=2, col sep = comma] {plots/Correctness/precision_cora_GCN.txt};\label{pre_cora_GCN_grad}
			
			\addplot [color=green, mark=diamond*, line width = 1pt] table [x index=0, y index=3, col sep = comma] {plots/Correctness/precision_cora_GCN.txt};\label{pre_cora_GCN_gradinput}
			
			\addplot [color=brown, mark=triangle*, line width = 1pt] table [x index=0, y index=4, col sep = comma] {plots/Correctness/precision_cora_GCN.txt};\label{pre_cora_GCN_smoothgrad}
			
			\addplot [color=blue, mark=square, line width = 1pt] table [x index=0, y index=5, col sep = comma] {plots/Correctness/precision_cora_GCN.txt};\label{pre_cora_GCN_ig}
			
		\end{axis}
	\end{tikzpicture}
        \end{subfigure}
	    \hfill
	    \begin{subfigure}{0.5\textwidth}
            \begin{tikzpicture}
		\begin{axis}[
			width  = \textwidth,
			height = 0.8\textwidth,
			major x tick style = transparent,
			grid = major,
		    grid style = {dashed, gray!20},
			ylabel = {Recall},
			title={GCN},
			symbolic x coords={5, 10, 15, 20, 25},
			xtick = data,
			enlarge x limits=0.25,
			]
			\addplot [color=black, mark=*, line width = 1pt] table [x index=0, y index=1, col sep = comma] {plots/Correctness/recall_cora_GCN.txt};\label{rec_cora_GCN_gnne}
			
			\addplot [color=red, mark=square*, line width = 1pt] table [x index=0, y index=2, col sep = comma] {plots/Correctness/recall_cora_GCN.txt};\label{rec_cora_GCN_grad}
			
			\addplot [color=green, mark=diamond*, line width = 1pt] table [x index=0, y index=3, col sep = comma] {plots/Correctness/recall_cora_GCN.txt};\label{rec_cora_GCN_gradinput}
			
			\addplot [color=brown, mark=triangle*, line width = 1pt] table [x index=0, y index=4, col sep = comma] {plots/Correctness/recall_cora_GCN.txt};\label{rec_cora_GCN_smoothgrad}
			
			\addplot [color=blue, mark=square, line width = 1pt] table [x index=0, y index=5, col sep = comma] {plots/Correctness/recall_cora_GCN.txt};\label{rec_cora_GCN_ig}

		\end{axis}
	\end{tikzpicture}
        \end{subfigure}
 \begin{subfigure}{0.5\textwidth}
            \begin{tikzpicture}
		\begin{axis}[
			width  = \textwidth,
			height = 0.8\textwidth,
			major x tick style = transparent,
			grid = major,
		    grid style = {dashed, gray!20},
			ylabel = {Precision},
			title={GAT},
			symbolic x coords={5, 10, 15, 20, 25},
			xtick = data,
			enlarge x limits=0.25,
			]
			
			\addplot [color=black, mark=*, line width = 1pt] table [x index=0, y index=1, col sep = comma] {plots/Correctness/precision_cora_GAT.txt};\label{pre_cora_GAT_gnne}
			
			\addplot [color=red, mark=square*, line width = 1pt] table [x index=0, y index=2, col sep = comma] {plots/Correctness/precision_cora_GAT.txt};\label{pre_cora_GAT_grad}
			
			\addplot [color=green, mark=diamond*, line width = 1pt] table [x index=0, y index=3, col sep = comma] {plots/Correctness/precision_cora_GAT.txt};\label{pre_cora_GAT_gradinput}
			
			\addplot [color=brown, mark=triangle*, line width = 1pt] table [x index=0, y index=4, col sep = comma] {plots/Correctness/precision_cora_GAT.txt};\label{pre_cora_GAT_smoothgrad}
			
			\addplot [color=blue, mark=square, line width = 1pt] table [x index=0, y index=5, col sep = comma] {plots/Correctness/precision_cora_GAT.txt};\label{pre_cora_GAT_ig}

		\end{axis}
	\end{tikzpicture}
        \end{subfigure}    
\begin{subfigure}{0.5\textwidth}
            \begin{tikzpicture}
		\begin{axis}[
			width  = \textwidth,
			height = 0.8\textwidth,
			major x tick style = transparent,
			grid = major,
		    grid style = {dashed, gray!20},
			ylabel = {Recall},
			title={GAT},
			symbolic x coords={5, 10, 15, 20, 25},
			xtick = data,
			enlarge x limits=0.25,
			]
			\addplot [color=black, mark=*, line width = 1pt] table [x index=0, y index=1, col sep = comma] {plots/Correctness/recall_cora_GAT.txt};\label{rec_cora_GAT_gnne}
			
			\addplot [color=red, mark=square*, line width = 1pt] table [x index=0, y index=2, col sep = comma] {plots/Correctness/recall_cora_GAT.txt};\label{rec_cora_GAT_grad}
			
			\addplot [color=green, mark=diamond*, line width = 1pt] table [x index=0, y index=3, col sep = comma] {plots/Correctness/recall_cora_GAT.txt};\label{rec_cora_GAT_gradinput}
			
			\addplot [color=brown, mark=triangle*, line width = 1pt] table [x index=0, y index=4, col sep = comma] {plots/Correctness/recall_cora_GAT.txt};\label{rec_cora_GAT_smoothgrad}
			
			\addplot [color=blue, mark=square, line width = 1pt] table [x index=0, y index=5, col sep = comma] {plots/Correctness/recall_cora_GAT.txt};\label{rec_cora_GAT_ig}
			
		\end{axis}
	\end{tikzpicture}
        \end{subfigure}      
\begin{subfigure}{0.5\textwidth}
            \begin{tikzpicture}
		\begin{axis}[
			width  = \textwidth,
			height = 0.8\textwidth,
			major x tick style = transparent,
			grid = major,
		    grid style = {dashed, gray!20},
			ylabel = {Precision},
			title={GIN},
			symbolic x coords={5, 10, 15, 20, 25},
			xtick = data,
			enlarge x limits=0.25,
			]
			
			\addplot [color=black, mark=*, line width = 1pt] table [x index=0, y index=1, col sep = comma] {plots/Correctness/precision_cora_GIN.txt};\label{pre_cora_GIN_gnne}
			
			\addplot [color=red, mark=square*, line width = 1pt] table [x index=0, y index=2, col sep = comma] {plots/Correctness/precision_cora_GIN.txt};\label{pre_cora_GIN_grad}
			
			\addplot [color=green, mark=diamond*, line width = 1pt] table [x index=0, y index=3, col sep = comma] {plots/Correctness/precision_cora_GIN.txt};\label{pre_cora_GIN_gradinput}
			
			\addplot [color=brown, mark=triangle*, line width = 1pt] table [x index=0, y index=4, col sep = comma] {plots/Correctness/precision_cora_GIN.txt};\label{pre_cora_GIN_smoothgrad}
			
			\addplot [color=blue, mark=square, line width = 1pt] table [x index=0, y index=5, col sep = comma] {plots/Correctness/precision_cora_GIN.txt};\label{pre_cora_GIN_ig}
			
		\end{axis}
	\end{tikzpicture}
        \end{subfigure}         
\begin{subfigure}{0.5\textwidth}
            \begin{tikzpicture}
		\begin{axis}[
			width  = \textwidth,
			height = 0.8\textwidth,
			major x tick style = transparent,
			grid = major,
		    grid style = {dashed, gray!20},
			ylabel = {Recall},
			title={GIN},
			symbolic x coords={5, 10, 15, 20, 25},
			xtick = data,
			enlarge x limits=0.25,
			]
			\addplot [color=black, mark=*, line width = 1pt] table [x index=0, y index=1, col sep = comma] {plots/Correctness/recall_cora_GIN.txt};\label{rec_cora_GIN_gnne}
			
			\addplot [color=red, mark=square*, line width = 1pt] table [x index=0, y index=2, col sep = comma] {plots/Correctness/recall_cora_GIN.txt};\label{rec_cora_GIN_grad}
			
			\addplot [color=green, mark=diamond*, line width = 1pt] table [x index=0, y index=3, col sep = comma] {plots/Correctness/recall_cora_GIN.txt};\label{rec_cora_GIN_gradinput}
			
			\addplot [color=brown, mark=triangle*, line width = 1pt] table [x index=0, y index=4, col sep = comma] {plots/Correctness/recall_cora_GIN.txt};\label{rec_cora_GIN_smoothgrad}
			
			\addplot [color=blue, mark=square, line width = 1pt] table [x index=0, y index=5, col sep = comma] {plots/Correctness/recall_cora_GIN.txt};\label{rec_cora_GIN_ig}
		\end{axis}
	\end{tikzpicture}
        \end{subfigure}         
\begin{subfigure}{0.5\textwidth}
            \begin{tikzpicture}
		\begin{axis}[
			width  = \textwidth,
			height = 0.8\textwidth,
			major x tick style = transparent,
			grid = major,
		    grid style = {dashed, gray!20},
			xlabel = {topk},
			ylabel = {Precision},
			title={APPNP},
			symbolic x coords={5, 10, 15, 20, 25},
			xtick = data,
			enlarge x limits=0.25,
			]

			\addplot [color=black, mark=*, line width = 1pt] table [x index=0, y index=1, col sep = comma] {plots/Correctness/precision_cora_APPNP.txt};\label{pre_cora_APPNP_gnne}
			
			\addplot [color=red, mark=square*, line width = 1pt] table [x index=0, y index=2, col sep = comma] {plots/Correctness/precision_cora_APPNP.txt};\label{pre_cora_APPNP_grad}
			
			\addplot [color=green, mark=diamond*, line width = 1pt] table [x index=0, y index=3, col sep = comma] {plots/Correctness/precision_cora_APPNP.txt};\label{pre_cora_APPNP_gradinput}
			
			\addplot [color=brown, mark=triangle*, line width = 1pt] table [x index=0, y index=4, col sep = comma] {plots/Correctness/precision_cora_APPNP.txt};\label{pre_cora_APPNP_smoothgrad}
			
			\addplot [color=blue, mark=square, line width = 1pt] table [x index=0, y index=5, col sep = comma] {plots/Correctness/precision_cora_APPNP.txt};\label{pre_cora_APPNP_ig}

		\end{axis}
	\end{tikzpicture}
        \end{subfigure}         
\begin{subfigure}{0.5\textwidth}
            \begin{tikzpicture}
		\begin{axis}[
			width  = \textwidth,
			height = 0.8\textwidth,
			major x tick style = transparent,
			grid = major,
		    grid style = {dashed, gray!20},
			xlabel = {topk},
			ylabel = {Recall},
			title={APPNP},
			symbolic x coords={5, 10, 15, 20, 25},
			xtick = data,
			enlarge x limits=0.25,
			]
			\addplot [color=black, mark=*, line width = 1pt] table [x index=0, y index=1, col sep = comma] {plots/Correctness/recall_cora_APPNP.txt};\label{rec_cora_APPNP_gnne}
			
			\addplot [color=red, mark=square*, line width = 1pt] table [x index=0, y index=2, col sep = comma] {plots/Correctness/recall_cora_APPNP.txt};\label{rec_cora_APPNP_grad}
			
			\addplot [color=green, mark=diamond*, line width = 1pt] table [x index=0, y index=3, col sep = comma] {plots/Correctness/recall_cora_APPNP.txt};\label{rec_cora_APPNP_gradinput}
			
			\addplot [color=brown, mark=triangle*, line width = 1pt] table [x index=0, y index=4, col sep = comma] {plots/Correctness/recall_cora_APPNP.txt};\label{rec_cora_APPNP_smoothgrad}
			
			\addplot [color=blue, mark=square, line width = 1pt] table [x index=0, y index=5, col sep = comma] {plots/Correctness/recall_cora_APPNP.txt};\label{rec_cora_APPNP_ig}

		\end{axis}
	\end{tikzpicture}
        \end{subfigure}         
	\caption{Correctness for \cora. GNNExp~\ref{pre_cora_GCN_gnne}~~Grad~\ref{pre_cora_GCN_grad}~~GradInput~\ref{pre_cora_GCN_gradinput}~~SmoothGrad~\ref{pre_cora_GCN_smoothgrad}~~IG~\ref{pre_cora_GCN_ig}}
    \label{fig:correctness_cora_topk}
\end{figure}

\begin{figure}[]

        \begin{subfigure}{0.5\columnwidth}
            \begin{tikzpicture}
		\begin{axis}[
			width  = \textwidth,
			height = 0.8\textwidth,
			major x tick style = transparent,
			grid = major,
		    grid style = {dashed, gray!20},
			ylabel = {Precision},
			title={GCN},
			symbolic x coords={5, 10, 15, 20, 25},
			xtick = data,
			enlarge x limits=0.25,
			]
			\addplot [color=black, mark=*, line width = 1pt] table [x index=0, y index=1, col sep = comma] {plots/Correctness/precision_citeseer_GCN.txt};\label{pre_citeseer_GCN_gnne}
			
			\addplot [color=red, mark=square*, line width = 1pt] table [x index=0, y index=2, col sep = comma] {plots/Correctness/precision_citeseer_GCN.txt};\label{pre_citeseer_GCN_grad}
			
			\addplot [color=green, mark=diamond*, line width = 1pt] table [x index=0, y index=3, col sep = comma] {plots/Correctness/precision_citeseer_GCN.txt};\label{pre_citeseer_GCN_gradinput}
			
			\addplot [color=brown, mark=triangle*, line width = 1pt] table [x index=0, y index=4, col sep = comma] {plots/Correctness/precision_citeseer_GCN.txt};\label{pre_citeseer_GCN_smoothgrad}
			
			\addplot [color=blue, mark=square, line width = 1pt] table [x index=0, y index=5, col sep = comma] {plots/Correctness/precision_citeseer_GCN.txt};\label{pre_citeseer_GCN_ig}
			
		\end{axis}
	\end{tikzpicture}
        \end{subfigure}
	    \hfill
	    \begin{subfigure}{0.5\textwidth}
            \begin{tikzpicture}
		\begin{axis}[
			width  = \textwidth,
			height = 0.8\textwidth,
			major x tick style = transparent,
			grid = major,
		    grid style = {dashed, gray!20},
			ylabel = {Recall},
			title={GCN},
			symbolic x coords={5, 10, 15, 20, 25},
			xtick = data,
			enlarge x limits=0.25,
			]
			\addplot [color=black, mark=*, line width = 1pt] table [x index=0, y index=1, col sep = comma] {plots/Correctness/recall_citeseer_GCN.txt};\label{rec_citeseer_GCN_gnne}
			
			\addplot [color=red, mark=square*, line width = 1pt] table [x index=0, y index=2, col sep = comma] {plots/Correctness/recall_citeseer_GCN.txt};\label{rec_citeseer_GCN_grad}
			
			\addplot [color=green, mark=diamond*, line width = 1pt] table [x index=0, y index=3, col sep = comma] {plots/Correctness/recall_citeseer_GCN.txt};\label{rec_citeseer_GCN_gradinput}
			
			\addplot [color=brown, mark=triangle*, line width = 1pt] table [x index=0, y index=4, col sep = comma] {plots/Correctness/recall_citeseer_GCN.txt};\label{rec_citeseer_GCN_smoothgrad}
			
			\addplot [color=blue, mark=square, line width = 1pt] table [x index=0, y index=5, col sep = comma] {plots/Correctness/recall_citeseer_GCN.txt};\label{rec_citeseer_GCN_ig}

		\end{axis}
	\end{tikzpicture}
        \end{subfigure}
 \begin{subfigure}{0.5\textwidth}
            \begin{tikzpicture}
		\begin{axis}[
			width  = \textwidth,
			height = 0.8\textwidth,
			major x tick style = transparent,
			grid = major,
		    grid style = {dashed, gray!20},
			ylabel = {Precision},
			title={GAT},
			symbolic x coords={5, 10, 15, 20, 25},
			xtick = data,
			enlarge x limits=0.25,
			]
			
			\addplot [color=black, mark=*, line width = 1pt] table [x index=0, y index=1, col sep = comma] {plots/Correctness/precision_citeseer_GAT.txt};\label{pre_citeseer_GAT_gnne}
			
			\addplot [color=red, mark=square*, line width = 1pt] table [x index=0, y index=2, col sep = comma] {plots/Correctness/precision_citeseer_GAT.txt};\label{pre_citeseer_GAT_grad}
			
			\addplot [color=green, mark=diamond*, line width = 1pt] table [x index=0, y index=3, col sep = comma] {plots/Correctness/precision_citeseer_GAT.txt};\label{pre_citeseer_GAT_gradinput}
			
			\addplot [color=brown, mark=triangle*, line width = 1pt] table [x index=0, y index=4, col sep = comma] {plots/Correctness/precision_citeseer_GAT.txt};\label{pre_citeseer_GAT_smoothgrad}
			
			\addplot [color=blue, mark=square, line width = 1pt] table [x index=0, y index=5, col sep = comma] {plots/Correctness/precision_citeseer_GAT.txt};\label{pre_citeseer_GAT_ig}

		\end{axis}
	\end{tikzpicture}
        \end{subfigure}    
\begin{subfigure}{0.5\textwidth}
            \begin{tikzpicture}
		\begin{axis}[
			width  = \textwidth,
			height = 0.8\textwidth,
			major x tick style = transparent,
			grid = major,
		    grid style = {dashed, gray!20},
			ylabel = {Recall},
			title={GAT},
			symbolic x coords={5, 10, 15, 20, 25},
			xtick = data,
			enlarge x limits=0.25,
			]
			\addplot [color=black, mark=*, line width = 1pt] table [x index=0, y index=1, col sep = comma] {plots/Correctness/recall_citeseer_GAT.txt};\label{rec_citeseer_GAT_gnne}
			
			\addplot [color=red, mark=square*, line width = 1pt] table [x index=0, y index=2, col sep = comma] {plots/Correctness/recall_citeseer_GAT.txt};\label{rec_citeseer_GAT_grad}
			
			\addplot [color=green, mark=diamond*, line width = 1pt] table [x index=0, y index=3, col sep = comma] {plots/Correctness/recall_citeseer_GAT.txt};\label{rec_citeseer_GAT_gradinput}
			
			\addplot [color=brown, mark=triangle*, line width = 1pt] table [x index=0, y index=4, col sep = comma] {plots/Correctness/recall_citeseer_GAT.txt};\label{rec_citeseer_GAT_smoothgrad}
			
			\addplot [color=blue, mark=square, line width = 1pt] table [x index=0, y index=5, col sep = comma] {plots/Correctness/recall_citeseer_GAT.txt};\label{rec_citeseer_GAT_ig}
			
		\end{axis}
	\end{tikzpicture}
        \end{subfigure}      
\begin{subfigure}{0.5\textwidth}
            \begin{tikzpicture}
		\begin{axis}[
			width  = \textwidth,
			height = 0.8\textwidth,
			major x tick style = transparent,
			grid = major,
		    grid style = {dashed, gray!20},
			ylabel = {Precision},
			title={GIN},
			symbolic x coords={5, 10, 15, 20, 25},
			xtick = data,
			enlarge x limits=0.25,
			]
			
			\addplot [color=black, mark=*, line width = 1pt] table [x index=0, y index=1, col sep = comma] {plots/Correctness/precision_citeseer_GIN.txt};\label{pre_citeseer_GIN_gnne}
			
			\addplot [color=red, mark=square*, line width = 1pt] table [x index=0, y index=2, col sep = comma] {plots/Correctness/precision_citeseer_GIN.txt};\label{pre_citeseer_GIN_grad}
			
			\addplot [color=green, mark=diamond*, line width = 1pt] table [x index=0, y index=3, col sep = comma] {plots/Correctness/precision_citeseer_GIN.txt};\label{pre_citeseer_GIN_gradinput}
			
			\addplot [color=brown, mark=triangle*, line width = 1pt] table [x index=0, y index=4, col sep = comma] {plots/Correctness/precision_citeseer_GIN.txt};\label{pre_citeseer_GIN_smoothgrad}
			
			\addplot [color=blue, mark=square, line width = 1pt] table [x index=0, y index=5, col sep = comma] {plots/Correctness/precision_citeseer_GIN.txt};\label{pre_citeseer_GIN_ig}
			
		\end{axis}
	\end{tikzpicture}
        \end{subfigure}         
\begin{subfigure}{0.5\textwidth}
            \begin{tikzpicture}
		\begin{axis}[
			width  = \textwidth,
			height = 0.8\textwidth,
			major x tick style = transparent,
			grid = major,
		    grid style = {dashed, gray!20},
			ylabel = {Recall},
			title={GIN},
			symbolic x coords={5, 10, 15, 20, 25},
			xtick = data,
			enlarge x limits=0.25,
			]
			\addplot [color=black, mark=*, line width = 1pt] table [x index=0, y index=1, col sep = comma] {plots/Correctness/recall_citeseer_GIN.txt};\label{rec_citeseer_GIN_gnne}
			
			\addplot [color=red, mark=square*, line width = 1pt] table [x index=0, y index=2, col sep = comma] {plots/Correctness/recall_citeseer_GIN.txt};\label{rec_citeseer_GIN_grad}
			
			\addplot [color=green, mark=diamond*, line width = 1pt] table [x index=0, y index=3, col sep = comma] {plots/Correctness/recall_citeseer_GIN.txt};\label{rec_citeseer_GIN_gradinput}
			
			\addplot [color=brown, mark=triangle*, line width = 1pt] table [x index=0, y index=4, col sep = comma] {plots/Correctness/recall_citeseer_GIN.txt};\label{rec_citeseer_GIN_smoothgrad}
			
			\addplot [color=blue, mark=square, line width = 1pt] table [x index=0, y index=5, col sep = comma] {plots/Correctness/recall_citeseer_GIN.txt};\label{rec_citeseer_GIN_ig}
		\end{axis}
	\end{tikzpicture}
        \end{subfigure}         
\begin{subfigure}{0.5\textwidth}
            \begin{tikzpicture}
		\begin{axis}[
			width  = \textwidth,
			height = 0.8\textwidth,
			major x tick style = transparent,
			grid = major,
		    grid style = {dashed, gray!20},
			xlabel = {topk},
			ylabel = {Precision},
			title={APPNP},
			symbolic x coords={5, 10, 15, 20, 25},
			xtick = data,
			enlarge x limits=0.25,
			]
			
			\addplot [color=black, mark=*, line width = 1pt] table [x index=0, y index=1, col sep = comma] {plots/Correctness/precision_citeseer_APPNP.txt};\label{pre_citeseer_APPNP_gnne}
			
			\addplot [color=red, mark=square*, line width = 1pt] table [x index=0, y index=2, col sep = comma] {plots/Correctness/precision_citeseer_APPNP.txt};\label{pre_citeseer_APPNP_grad}
			
			\addplot [color=green, mark=diamond*, line width = 1pt] table [x index=0, y index=3, col sep = comma] {plots/Correctness/precision_citeseer_APPNP.txt};\label{pre_citeseer_APPNP_gradinput}
			
			\addplot [color=brown, mark=triangle*, line width = 1pt] table [x index=0, y index=4, col sep = comma] {plots/Correctness/precision_citeseer_APPNP.txt};\label{pre_citeseer_APPNP_smoothgrad}
			
			\addplot [color=blue, mark=square, line width = 1pt] table [x index=0, y index=5, col sep = comma] {plots/Correctness/precision_citeseer_APPNP.txt};\label{pre_citeseer_APPNP_ig}

		\end{axis}
	\end{tikzpicture}
        \end{subfigure}         
\begin{subfigure}{0.5\textwidth}
            \begin{tikzpicture}
		\begin{axis}[
			width  = \textwidth,
			height = 0.8\textwidth,
			major x tick style = transparent,
			grid = major,
		    grid style = {dashed, gray!20},
			xlabel = {topk},
			ylabel = {Recall},
			title={APPNP},
			symbolic x coords={5, 10, 15, 20, 25},
			xtick = data,
			enlarge x limits=0.25,
			]
			\addplot [color=black, mark=*, line width = 1pt] table [x index=0, y index=1, col sep = comma] {plots/Correctness/recall_citeseer_APPNP.txt};\label{rec_citeseer_APPNP_gnne}
			
			\addplot [color=red, mark=square*, line width = 1pt] table [x index=0, y index=2, col sep = comma] {plots/Correctness/recall_citeseer_APPNP.txt};\label{rec_citeseer_APPNP_grad}
			
			\addplot [color=green, mark=diamond*, line width = 1pt] table [x index=0, y index=3, col sep = comma] {plots/Correctness/recall_citeseer_APPNP.txt};\label{rec_citeseer_APPNP_gradinput}
			
			\addplot [color=brown, mark=triangle*, line width = 1pt] table [x index=0, y index=4, col sep = comma] {plots/Correctness/recall_citeseer_APPNP.txt};\label{rec_citeseer_APPNP_smoothgrad}
			
			\addplot [color=blue, mark=square, line width = 1pt] table [x index=0, y index=5, col sep = comma] {plots/Correctness/recall_citeseer_APPNP.txt};\label{rec_citeseer_APPNP_ig}

		\end{axis}
	\end{tikzpicture}
        \end{subfigure}         
	\caption{Correctness for \citeseer. GNNExp~\ref{pre_citeseer_GCN_gnne}~~Grad~\ref{pre_citeseer_GCN_grad}~~GradInput~\ref{pre_citeseer_GCN_gradinput}~~SmoothGrad~\ref{pre_citeseer_GCN_smoothgrad}~~IG~\ref{pre_citeseer_GCN_ig}}
    \label{fig:correctness_citeseer_topk}
	
\end{figure}
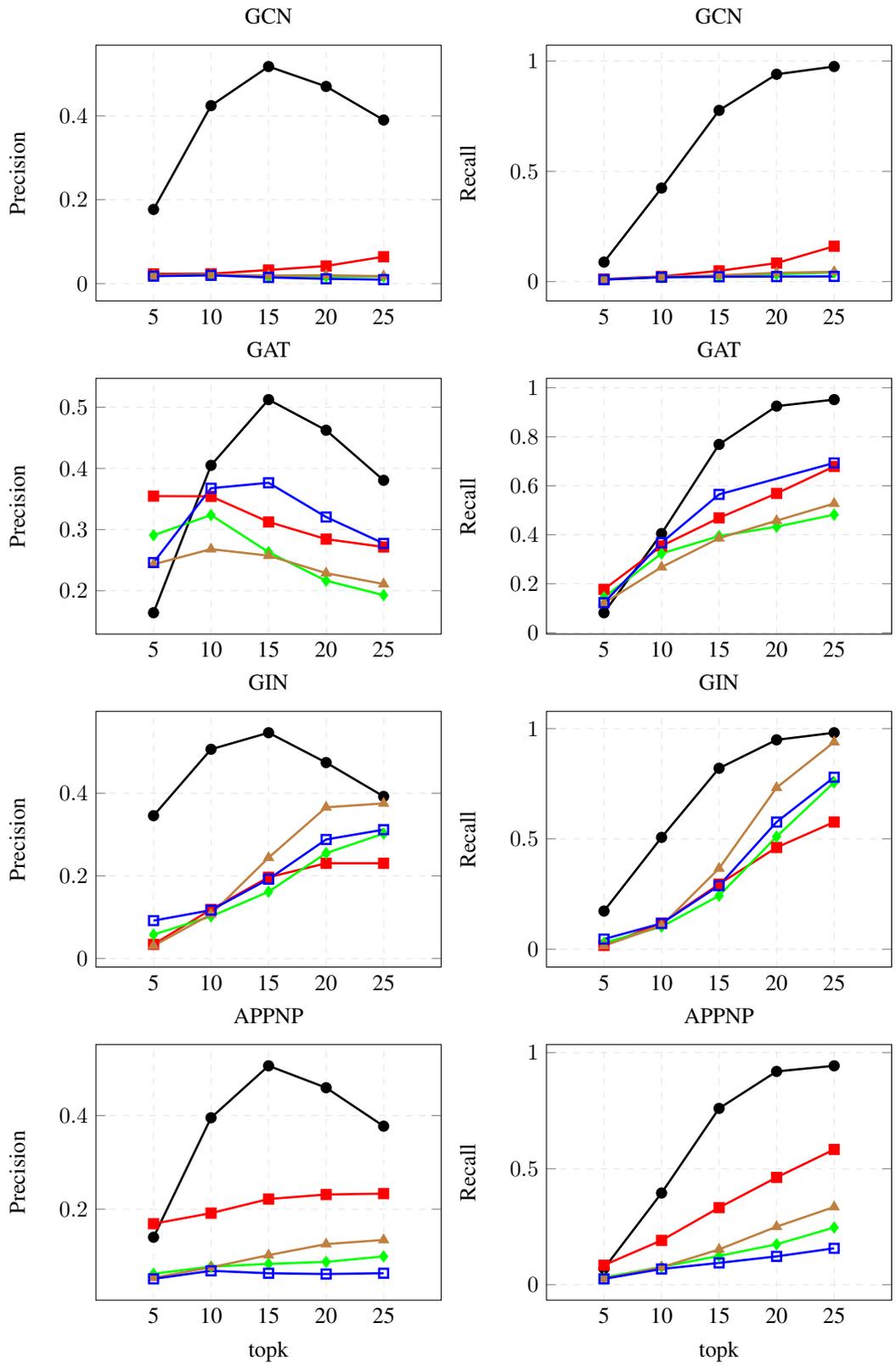

In Figure~\ref{fig:movie_review_topk}, we report token level macro-F1 (ma-F1) and micro-F1 (mi-F1) scores with different $topk\in[5,10,15,20,25]$ with different GNN models. Figure~\ref{fig:correctness_cora_topk} and~\ref{fig:correctness_citeseer_topk} show the correctness for \cora and \citeseer respectively with different \textit{topk}.

\section{Limitations}
\label{sec:limitation}
The major limitation of this work is that \ourbench only focuses on the \textit{post-hoc explanation} approaches. This work does not cover \textit{interpretable-by-design} and \textit{counter factual based explanation} methods. 
Though we tried our best to use datasets with varying graph properties and distributions, we believe this benchmark has a large scope to expand to multiple graph datasets with varying graph properties. 
We strive to increase the number of more varied datasets in \ourbench in future.

\end{document}